%% file: arxiv_main.tex
\documentclass[]{dreamvu_preamble/dreamvu}
\pdfoutput=1

\usepackage{verbatim}
\usepackage{amsmath,amsfonts,amssymb}
\usepackage{xspace}
\usepackage{url}
\usepackage{array}
\usepackage{tabularx}
\usepackage{adjustbox}
\usepackage{wrapfig}
\usepackage{lmodern}
\usepackage{enumitem}
\usepackage{cleveref}
\usepackage{comment}
\usepackage{float}
\usepackage{tikz}
\usetikzlibrary{arrows.meta}
\usepackage{pgfplots}
\pgfplotsset{compat=1.18}
\usepackage{xcolor}
\definecolor{darkgreen}{rgb}{0.0, 0.5, 0.0}

\newcommand{\nbf}[1]{\noindent\textbf{#1.}~}

\newtoggle{arxiv}
\toggletrue{arxiv}

\title{PRISM: A Multi-View Multi-Capability Retail Video Dataset for Embodied Vision-Language Models}

\author{DreamVu\protect\footnotemark}

\abstract{\input{00_abstract_shortened}
}

\date{\today}

\begin{document}

\maketitle

\footnotetext{A detailed list of contributors and acknowledgments can be found in Appendix \hyperref[sec:contribution]{C} of this paper}

\begin{figure*}[h!]
    \centering
    \includegraphics[width=0.9\textwidth]{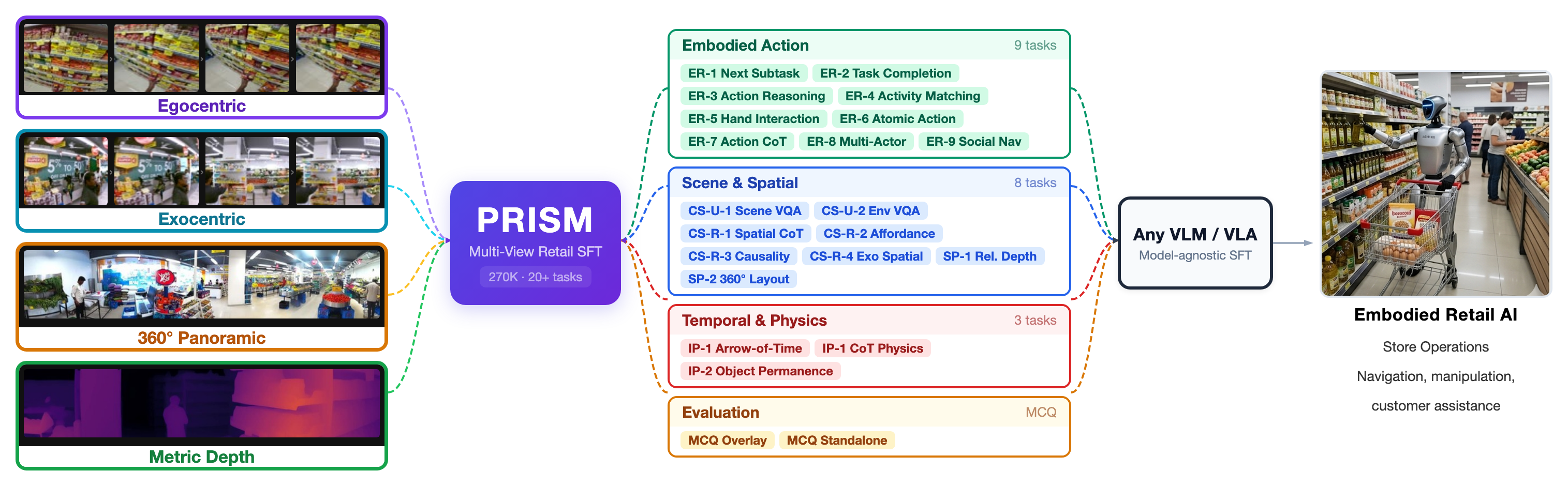}
    \vspace{-6pt}
    \caption{\textbf{PRISM} captures multi-view retail video from four synchronized modalities - egocentric, exocentric, 360\textdegree{} panoramic and depth - and structures 270K video SFT samples across 20+ task types organized into four capability dimensions. All capabilities feed into a model-agnostic fine-tuning format compatible with any VLM or VLA, producing embodied agents for real-world retail deployment.}
    \label{fig:teaser}
\end{figure*}

\input{01_intro}

\input{02_related}
\input{03_method}
\input{04_experiments}

\input{05_results}
\input{10_conclusion}

\clearpage
\bibliographystyle{plainnat}
\bibliography{11_references}

\clearpage
\beginappendix

\input{12_appendix}

\end{document}

%% file: 00_abstract_shortened.tex
A critical gap exists between the general-purpose visual understanding of state-of-the-art physical AI models and the specialized perceptual demands of structured real-world deployment environments. We present \textbf{PRISM}, a 270K-sample multi-view video supervised fine-tuning (SFT) corpus for embodied vision-language-models (VLMs) in real-world retail environments. PRISM is motivated by a simple observation - physical AI systems fail not because of poor visual recognition, but because they do not understand space, physical dynamics and embodied action well enough to operate reliably in the world.

To this end, PRISM is grounded in a novel three-dimensional knowledge ontology that spans spatial knowledge, temporal and physical knowledge, and embodied action knowledge. It covers 20+ capability probes across four evaluation dimensions - Embodied Reasoning (ER), Common Sense (CS), Spatial Perception (SP), and Intuitive Physics (IP), and to our knowledge, PRISM is the first dataset to instantiate all three knowledge dimensions within a single real-world deployment domain. The corpus captures data from \textbf{egocentric}, \textbf{exocentric} and \textbf{360\textdegree{}} viewpoints across five supermarket locations and includes open-ended, chain-of-thought, and multiple-choice supervision. At 4\,fps, PRISM spans $\sim$11.8M video frames and $\sim$730M tokens, placing it among the largest domain-specific video SFT corpora.

Fine-tuning on PRISM reduces the error rate across all 20+  probes by \textbf{\textit{66.6\%}} over pre-trained baseline, with significant gains in embodied action understanding where the accuracy improves by \textbf{\textit{36.4\%}}. Our results suggest that ontology-structured, domain specific SFT can meaningfully strengthen embodied VLMs for real-world settings. The PRISM dataset and more details are available at \url{https://dreamvu.ai/prism}

%% file: 01_intro.tex
\section{Introduction}
\label{sec:intro}

Consider a robot that navigates a grocery aisle. It must estimate whether a shelf height is within reach, predict what a partially occluded worker will do next, determine whether its current task is complete, and interpret same scene from its body-mounted camera. Each of these requires a distinct perceptual capability- spatial depth reasoning, action anticipation, task completion verification, and cross-view understanding, that current general-purpose vision-language models do not reliably provide. This is not a failure of the underlying architectures. It is a failure of the training data.

Vision-language models (VLMs) are emerging as powerful backbones for embodied AI systems, enabling them to perceive, reason about, and act in physical environments through natural language inputs~\cite{cosmos_reason2, groot_n1, rt2}. Recent models such as Cosmos-Reason2~\cite{cosmos_reason2}, GR00T N1~\cite{groot_n1}, and $\pi_0$~\cite{pi0} demonstrate that multi-modal reasoning is increasingly essential for robotic manipulation, navigation, and task execution in unstructured real-world settings. However, a significant gap remains between the broad visual competence of these models and the specialized perceptual demands of specific deployment domains such as retail environments.

Fine-tuning VLMs for embodied applications requires carefully curated instruction-tuning datasets covering diverse perceptual capabilities. Although prior datasets for general video QA~\cite{videochatgpt, video_mme}, temporal reasoning~\cite{mvbench}, and physical understanding~\cite{physbench} have advanced open-domain evaluation, they do not adequately capture the range of knowledge required for physical AI in structured real-world environments. In particular, there is a scarcity of large-scale datasets targeting embodied perception across multiple knowledge dimensions in such environments. 
In our view, effective physical AI demands mastery across three fundamental and complementary forms of knowledge: spatial knowledge--the ability to perceive scene geometry, depth, and 3D layout; temporal and physical knowledge--the ability to reason about time, causality, and the laws governing object motion; and embodied action knowledge-the ability to anticipate actions, goals, task progress, and human-environment interaction. Existing datasets address at most one of these dimensions in any given domain, leaving a critical gap for real-world deployment.

Real-world retail spaces present several unique challenges: dense object arrangements, repetitive visual patterns at multiple scales, complex human-object interactions, and the need for 3D-aware spatial reasoning in cluttered aisles. These settings require models to understand not only what appears in a frame, but how actions unfold in space, how agents move through shared spaces, and how physical context constrains behavior. Virtually all existing embodied datasets are collected exclusively from egocentric (first-person, body mounted) cameras, introducing fundamental limitations that are rarely acknowledged. The narrow field-of-view of an egocentric camera captures only what the wearer attends to, leaving the broader scene structure--aisle geometry, shelf arrangements, navigation corridors, and spatial relationships between zones--entirely unobserved. Egocentric footage cannot capture other people in the scene except as fleeting peripheral appearances, rendering it unsuitable for multi-human activity recognition, social navigation reasoning, and any form of multi-actor trajectory understanding. Moreover, the constant motion of an egocentric camera introduces depth ambiguity and removes the stable reference frame for reliable 3D reasoning. In practice, real deployments must operate across both egocentric and exocentric views, yet existing datasets rarely provide the cross-view supervision needed to support such training.

In this paper, we introduce \textbf{PRISM}, a comprehensive multi-view video supervised fine-tuning (SFT) corpus for training embodied VLMs in retail environments. PRISM is structured around a novel knowledge ontology that interconnects the three core dimensions of physical AI - space, time and embodied action - within a single deployment domain (\cref{fig:teaser}). PRISM provides data for cross-view supervision that is captured from egocentric, exocentric and 360\textdegree{} viewpoints. It comprises of 270K instruction-tuning samples spanning 20+ capability probes across four evaluation dimensions - Embodied Reasoning, Common Sense, Spatial Perception, and Intuitive Physics - making it, to our knowledge, the first dataset to unify all three knowledge dimensions in a single real-world setting.

\input{figs/overview}

PRISM is also designed for breadth of supervision. It includes multiple SFT formats, such as open-ended understanding, chain-of-thought reasoning, and multiple-choice question answering, enabling the study of what models learn and also how training format shapes embodied reasoning. At 4\,fps with Qwen3VL tokenization, PRISM totals $\sim$11.8M video frames and $\sim$730M tokens ($\sim$703M visual + $\sim$27M text), placing it among the largest domain-specific SFT corpora for embodied VLMs. A comparison against Cosmos-Reason training data shows that \textbf{14 of the 20+ PRISM tasks are novel} - 6 are completely new with no Cosmos equivalent, 8 introduce new task types or methods within Cosmos categories,  three overlap conceptually but differ substantially in domain, and only two overlap completely. An overview of the PRISM pipeline is provided in \cref{fig:overview}

The capability probes in PRISM are constructed to reflect the diversity of reasoning required in real-world retail settings. Embodied Reasoning tasks include next-subtask prediction, goal-conditioned action reasoning, hand-interaction recognition, atomic action understanding from exocentric views, multi-actor scene understanding, social navigation reasoning, and exo-to-ego activity matching. Common-Sense and Spatial Perception tasks include scene description visual-question-answers (VQA), exocentric environment VQA, affordance reasoning, causality reasoning, depth-aware spatial reasoning, relative depth reasoning and 360\textdegree{} spatial layout understanding. Intuitive Physics tasks include arrow-of-time prediction from both ego and exo views, object permanence in real-world video and physics-grounded temporal reasoning. Together, these tasks provide a comprehensive training and evaluation basis for embodied perception across multiple viewpoints and knowledge types. 

A key design principle of PRISM is \textit{scalable construction at minimal cost}. PRISM is built using a hybrid annotation strategy that combines structured metadata extraction from robot episode logs, LLM-based QA generation via Gemini~\cite{gemini25}, (3)~physics-grounded video reasoning via Gemini Robotics ER 1.5~\cite{gemini_robotics}, depth-aware spatial analysis using DepthCrafter ~\cite{depthcrafter}, and fully self-supervised transformations requiring zero annotation. This approach yields diverse, high-quality training data for the entire dataset with minimal human effort and API usage.

Beyond introducing the dataset itself, we use PRISM to conduct a systematic fine-tuning study over an embodied VLM for real-world deployment. Our results demonstrate that ontology-structured, domain-specific SFT substantially improves embodied VLM performance in retail environments. Multi-domain training consistently outperforms single-domain baselines, exocentric supervision improves cross-view understanding without degrading egocentric capability, and richer reasoning chains provide substantial gains over template-based alternatives. Our experiments suggest that progress in embodied VLMs depends not only on the scale of the model, but also on how training data is structured across knowledge dimensions, viewpoints and reasoning formats. Our main contributions are summarized hereafter:

\nbf{Contributions}
\begin{enumerate}[leftmargin=*,nosep]
    \item We introduce \textbf{PRISM}, a 270K-sample multi-view video SFT corpus for embodied VLMs in retail environments, organized around a knowledge ontology spanning embodied reasoning, common sense, spatial perception, and intuitive physics.
    \item We present the first systematic study of \textbf{egocentric and exocentric data mixing} as a curriculum dimension for embodied VLM fine-tuning, including dedicated exocentric capabilities for action understanding, spatial reasoning, and multi-actor scene analysis. We demonstrate quantifiable gains in real-world retail scene understanding for embodied actions.
\end{enumerate}

By unifying spatial, temporal-physical, and embodied action supervision across egocentric and exocentric views, PRISM provides a more deployment-aligned training substrate for embodied VLMs. Beyond the dataset itself, our results highlight that progress in physical AI depends not only on model scale, but also on how training data is structured across viewpoints, capabilities, and reasoning formats

%% file: figs/overview.tex
\begin{figure*}[t]
    \centering
    \resizebox{\textwidth}{!}{%
    \begin{tikzpicture}[
        box/.style={draw, rounded corners=3pt, minimum height=0.9cm, minimum width=2.8cm, align=center, font=\small},
        domainbox/.style={draw, rounded corners=3pt, minimum height=0.9cm, minimum width=2.2cm, align=center, font=\small\bfseries, line width=1pt},
        taskbox/.style={draw, rounded corners=2pt, minimum height=0.7cm, minimum width=3.6cm, align=center, font=\scriptsize},
        exobox/.style={draw, dashed, rounded corners=2pt, minimum height=0.7cm, minimum width=3.6cm, align=center, font=\scriptsize},
        arrow/.style={-{Stealth[length=2.5mm]}, thick},
        label/.style={font=\scriptsize\itshape, text=gray},
    ]

    \node[box, fill=blue!8]  at (0,  0.5) (ego) {Egocentric Video\\};
    \node[box, fill=blue!18] at (0, -1.5) (exo) {Exocentric Video\\(+ 360\textdegree{})};

    \node[label] at (4.5, 3.5) {Annotation Pipeline};
    \node[box, fill=orange!10] at (4.5,  2.5) (meta)  {Metadata Extraction\\(Gemini Robotics ER 1.5)};
    \node[box, fill=orange!10] at (4.5,  0.5) (llm)   {LLM Generation\\(Gemini 2.5 Flash)};
    \node[box, fill=orange!10] at (4.5, -1.5) (depth) {Depth Analysis\\(DepthCrafter)};
    \node[box, fill=orange!10] at (4.5, -3.5) (self)  {Self-Supervised\\(no annotation)};

    \node[domainbox, fill=red!12]    at (9.0,  2.5) (er) {Embodied\\Reasoning (ER)};
    \node[domainbox, fill=green!12]  at (9.0, -0.5) (cs) {Common\\Sense (CS)};
    \node[domainbox, fill=teal!12]   at (9.0, -2.5) (sp) {Spatial\\Perception (SP)};
    \node[domainbox, fill=cyan!12]   at (9.0, -4.5) (ip) {Intuitive\\Physics (IP)};

    \node[taskbox, fill=red!5] at (14.0,  4.0) (er1)    {ER-1: Next Subtask (25K) \textit{ego}};
    \node[taskbox, fill=red!5] at (14.0,  3.0) (er5)    {ER-5: Hand Interaction (15K) \textit{ego}};
    \node[exobox,  fill=red!2] at (14.0,  2.0) (er6)    {ER-6: Atomic Action (20K) \textit{exo}};
    \node[exobox,  fill=red!2] at (14.0,  1.0) (er8)    {ER-8: Multi-Actor (8K) \textit{exo}};

    \node[taskbox, fill=green!5] at (14.0,  0.0) (csu1)  {CS-U-1: Scene VQA (12K) \textit{ego}};
    \node[taskbox, fill=green!5] at (14.0, -1.0) (csr1)  {CS-R-1: Spatial CoT (12K) \textit{ego}};

    \node[taskbox, fill=teal!5]  at (14.0, -2.0) (sp1)   {SP-1: Rel.\ Depth (10K) \textit{ego}};
    \node[exobox,  fill=teal!2]  at (14.0, -3.0) (sp2)   {SP-2: 360\textdegree{} Layout (2K) \textit{exo}};

    \node[taskbox, fill=cyan!5] at (14.0, -4.0) (ip1)    {IP-1: Arrow-of-Time (30K) \textit{ego+exo}};
    \node[taskbox, fill=cyan!5] at (14.0, -5.0) (ip1c)   {IP-1 CoT: Physics (48K) \textit{ego+exo}};
    \node[taskbox, fill=cyan!5] at (14.0, -6.0) (ip2)    {IP-2: Object Permanence (8K) \textit{ego}};

    \node[font=\scriptsize\bfseries, text=red!60, anchor=west] at (16.0,  2.5) {+5 more};
    \node[font=\scriptsize\bfseries, text=green!60!black, anchor=west] at (16.0, -0.5) {+4 more};

    \node[box, fill=purple!10, minimum width=3.2cm, minimum height=2.4cm]
        at (19.5, -1.0) (ft) {BF16 LoRA Fine-Tuning\\Cosmos-Reason2-2B\\270K samples\\(ego + exo)};


    \draw[arrow] (ego.east) -- ++(0.4,0) |- (meta.west);
    \draw[arrow] (ego.east) -- (llm.west);
    \draw[arrow] (exo.east) -- (depth.west);
    \draw[arrow] (exo.east) -- ++(0.4,0) |- (self.west);

    \draw[arrow] (meta.east)  -- ++(0.4,0) |- (er.west);

    \draw[thick] (llm.east) -- (6.8,0.5);
    \draw[thick] (6.8,2.5) -- (6.8,-2.5);

    \draw[arrow] (6.8,2.5) -- (er.west);
    \draw[arrow] (6.8,-0.5) -- (cs.west);
    \draw[arrow] (6.8,-2.5) -- (sp.west);

    \draw[arrow] (depth.east) -- ++(0.4,0) |- (sp.west);
    \draw[arrow] (self.east)  -- ++(0.4,0) |- (ip.west);

    \draw[thick] (10.8,  4.0) -- (10.8,  1.0);
    \draw[thick] (er.east)    -- (10.8,  2.5);
    \draw[arrow] (10.8,  4.0) -- (er1.west);
    \draw[arrow] (10.8,  3.0) -- (er5.west);
    \draw[arrow] (10.8,  2.0) -- (er6.west);
    \draw[arrow] (10.8,  1.0) -- (er8.west);

    \draw[thick] (10.8,  0.0) -- (10.8, -1.0);
    \draw[thick] (cs.east)    -- (10.8, -0.5);
    \draw[arrow] (10.8,  0.0) -- (csu1.west);
    \draw[arrow] (10.8, -1.0) -- (csr1.west);

    \draw[thick] (10.8, -2.0) -- (10.8, -3.0);
    \draw[thick] (sp.east)    -- (10.8, -2.5);
    \draw[arrow] (10.8, -2.0) -- (sp1.west);
    \draw[arrow] (10.8, -3.0) -- (sp2.west);

    \draw[thick] (10.8, -4.0) -- (10.8, -6.0);
    \draw[thick] (ip.east)    -- (10.8, -4.5);
    \draw[arrow] (10.8, -4.0) -- (ip1.west);
    \draw[arrow] (10.8, -5.0) -- (ip1c.west);
    \draw[arrow] (10.8, -6.0) -- (ip2.west);

    \draw[dashed, thick] (16.5,  4.0) -- (16.5, -6.0);
    \draw[dashed, thick] (er1.east)   -- (16.5,  4.0);
    \draw[dashed, thick] (er5.east)   -- (16.5,  3.0);
    \draw[dashed, thick] (er6.east)   -- (16.5,  2.0);
    \draw[dashed, thick] (er8.east)   -- (16.5,  1.0);
    \draw[dashed, thick] (csu1.east)  -- (16.5,  0.0);
    \draw[dashed, thick] (csr1.east)  -- (16.5, -1.0);
    \draw[dashed, thick] (sp1.east)   -- (16.5, -2.0);
    \draw[dashed, thick] (sp2.east)   -- (16.5, -3.0);
    \draw[dashed, thick] (ip1.east)   -- (16.5, -4.0);
    \draw[dashed, thick] (ip1c.east)  -- (16.5, -5.0);
    \draw[dashed, thick] (ip2.east)   -- (16.5, -6.0);
    \draw[arrow, dashed] (16.5, -1.0) -- (ft.west);

    \node[taskbox, fill=white, minimum width=2.0cm] at (14.0, -7.0) (leg1) {\textit{ego} task};
    \node[exobox,  fill=white, minimum width=2.0cm] at (16.6, -7.0) (leg2) {\textit{exo} task};
    \node[label] at (12.0, -7.0) {Legend:};

    \end{tikzpicture}%
    }
    \caption{\textbf{PRISM pipeline overview.} The overview of the PRISM pipeline is shown here. PRISM is built on egocentric and exocentric videos from real-world retail stores. Four annotation strategies - metadata extraction, LLM generation (Gemini 2.5 Flash), depth analysis (DepthCrafter), and self-supervised transformations - produce twenty tasks (full list of tasks is mentioned in section \cref{sec:method}) across four capability domains, totaling 270K instruction-tuning samples. Solid boxes denote egocentric tasks; dashed boxes denote exocentric tasks (\textit{exo}). Samples from PRISM are used for finetuning Cosmos-Reason2-2B via BF16 LoRA.}
    \label{fig:overview}
\end{figure*}
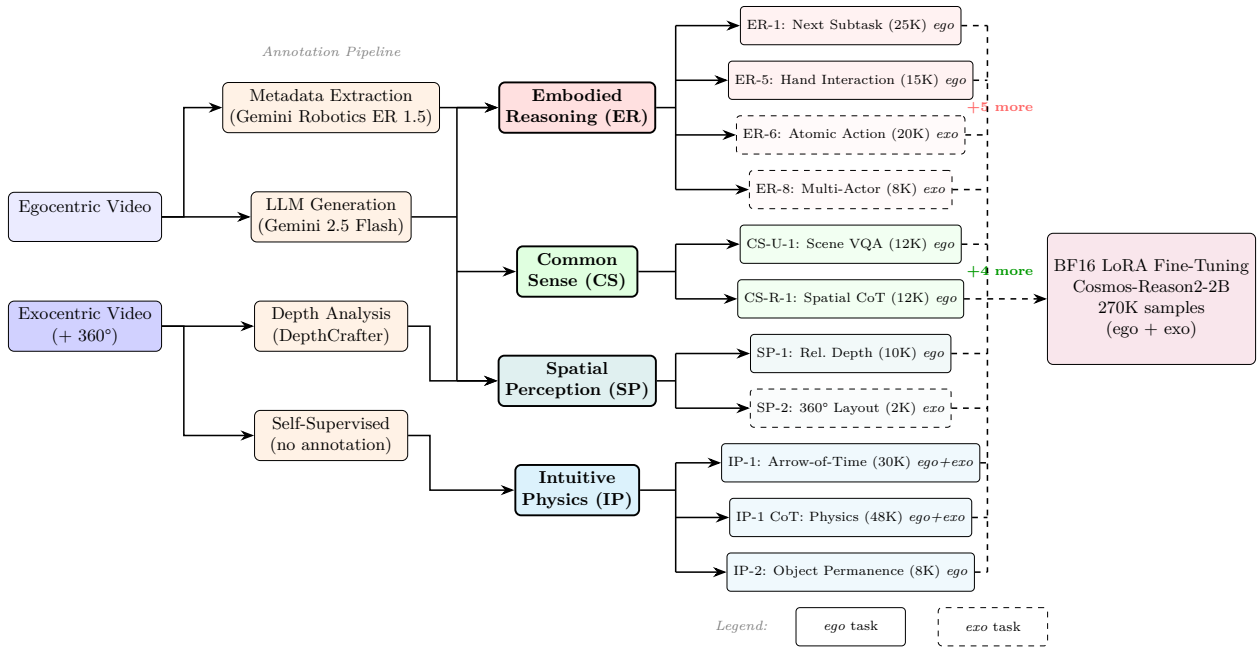

%% file: 02_related.tex
\section{Related Work}
\label{sec:related}

\nbf{Vision-Language Models for Embodied AI}
The integration of vision and language understanding has driven rapid progress in embodied AI. Models such as RT-2 ~\cite{rt2} demonstrated that VLMs can directly produce robotic actions, while PaLM-E ~\cite{palme} showed that large-scale multi-modal models can serve as general-purpose embodied reasoners. More recently, NVIDIA's Cosmos-Reason ~\cite{cosmos_reason1} and Cosmos-Reason2 ~\cite{cosmos_reason2} introduced physical AI-oriented VLMs trained on large-scale video corpora with chain-of-thought reasoning supervision, achieving strong performance on embodied reasoning benchmarks. GR00T N1 ~\cite{groot_n1} further illustrated how VLM backbones can be effectively integrated into humanoid robot control pipelines. Together, these systems suggest that progress in embodied AI increasingly depends not only on perception or language in isolation, but on models that can jointly interpret scenes, actions and spatio-temporal phyiscal context. 

\nbf{Knowledge ontology gap in existing embodied VLMs}
Despite this progress, existing embodied VLM training datasets remain uneven in their coverage of the knowledge required for deployment. A close examination of the Cosmos-Reason2 training taxonomy ~\cite{cosmos_reason1,cosmos_reason2} reveals that its coverage of PRISM's knowledge ontology is sparse: only 2 of PRISM's 23 capabilities are directly addressed in Cosmos's training data, while most are either absent entirely or introduced only in substantially different forms. In particular, PRISM contributes supervisio for cross-view activity matching, exocentric action understanding, multi-actor scene reasoning, retail-specific affordance understanding and panoramic spatial reasoning, all of which are largely missing from existing physical AI corpora. This gap is not only about the task count, but also about the deployment setting. Cosmos-Reason is trained primarily on general indoor, office, lab and structured-robot scenarios whereas PRISM targets a real-world retail environment. PRISM is therefore positioned not as a replacement for general embodied pretraining, but as a targeted fine-tuning resource that meaningfully fills domain gaps. A detailed quantitative comparison is provided in \cref{tab:cosmos_comparison}.

\input{tables/cosmos_comparison}

\nbf{Egocentric and Exocentric Video Understanding}
Egocentric video understanding has grown rapidly with large-scale datasets such as Ego4D ~\cite{ego4d}, Ego-Exo4D ~\cite{egoexo4d}, and Epic-Kitchens ~\cite{epic_kitchens}, enabling progress in activity recognition, hand-object interaction, anticipation, and cross-view learning. Ego-Exo4D ~\cite{egoexo4d} specifically highlights the complementarity of synchronized first- and third-person views for understanding human activity. However, these datasets focus on household, cooking, sports and other skilled activities rather than deployment oriented environments.

PRISM focuses on a coherent deployment domain rather than broad activity coverage. It addresses the underexplored domain of retail environments and, critically, pairs dynamic egocentric camera footage with static exocentric cameras, enabling the first systematic study of ego-exo data mixing for embodied VLM fine-tuning. The nature of the exocentric perspective differs fundamentally between existing datasets and PRISM. In Ego-Exo4D and similar corpora, the exocentric camera is a participant-following setup - a directed camera tracking the primary actor to capture their skilled performance. In PRISM, we use the \cite{DreamVuAlia} omnidirectional camera that observes the full environment without privileging any single actor. This distinction is consequential for embodied deployment. Scene-oriented exocentric views simultaneously capture multiple people, global aisle layout, crowd density, spatial relationships bentyen zones, and navigation-relevant scene structure - information that is central for a robot's situational awareness but structurally absent from other actor-following exocentric setups.

In the retail domain specifically, prior work such as the MERL Shopping Dataset ~\cite{merl_shopping} established fixed overhead-camera action localization (5 hand/arm action classes at supermarket shelves, 106 clips), and the RetailVision workshop series ~\cite{retailvision} has driven product recognition and in-store action localization challenges across six CVPR/ICCV editions. More recently, RetailAction ~\cite{retailaction} introduced multi-view spatio-temporal annotations of human--object interactions from 360\textdegree{} ceiling-mounted cameras across real-world convenience stores. However, these datasets remain purely exocentric, focus primarily on action localization or recognition, and do not provide instruction-tuning data for VLMs.

\nbf{Instruction-Tuning Datasets for VLMs}
The success of visual instruction-tuning ~\cite{llava} has led to a large ecosystem of multi-modal instruction datasets. LLaVA ~\cite{llava} pioneered GPT-4-generated visual conversations, while subsequent efforts such as Video-ChatGPT ~\cite{videochatgpt}, LLaVA-OneVision ~\cite{llava_onevision}, and InternVL ~\cite{internvl} scaled instruction data to millions of samples across diverse tasks. For video understanding specifically, VideoChat2 ~\cite{videochat2} and Video-LLaVA ~\cite{video_llava} introduced temporal reasoning instructions. PRISM differs from these datasets by focusing on a coherent domain and by structuring instruction data around an explicit knowledge ontology rather than open-domain data crawling. In this sense, PRISM is not only a dataset, but also a controlled corpus for studying how instruction composition shapess embodied VLM adaptation.

\nbf{Data-Efficient VLM Fine-Tuning}
Parameter-efficient fine-tuning methods such as LoRA ~\cite{lora} and QLoRA ~\cite{qlora} have made it practical to adapt large models on limited hardware. At the same time, several recent works have shown that data composition can matter as much as model architecture. VILA ~\cite{vila}, for example, emphasized the importanve of careful data mixing strategies for instruction-tuning for VLMs. Similarly, LLaVA-OneVision ~\cite{llava_onevision} showed that multi-stage training with task-specific data mixtures outperforms naive data concatenation. Our work complements these findings by providing fine-grained analysis of domain composition, task selection, annotation quality, and data-scale effects within a single controlled deployment setting.

\nbf{Comparison with Existing Embodied Datasets}
\cref{tab:dataset_comparison} positions PRISM against the three most closely
related embodied video datasets.
Ego4D ~\cite{ego4d} and Ego-Exo4D ~\cite{egoexo4d} are the dominant benchmarks
for egocentric and ego-exo video understanding, collectively spanning thousands
of hours of annotated footage across general daily-life and skilled-activity
domains. However, neither is designed as a VLM instruction-tuning corpus, neither provides
chain-of-thought supervision, and neither targets a specific deployment domain
such as retail. A deeper limitation for humanoid learning is that both datasets are organized around \textit{short, isolated activity clips} - atomic skills such as pouring, cutting, or dribbling - rather than complete, long-horizon \textit{tasks} in which an agent pursues a goal across a sequence of connected sub-goals while continuously moving through and reasoning about a real-world environment. Training on skill-level clips produces models that recognize gestures and actions in isolation but lack the world-knowledge required for goal-directed navigation, multi-step planning, and sequential task execution in deployment settings. 

PRISM is structured around complete retail \textit{episodes} - full workflows from store entry through item retrieval, shelf interaction, and checkout - each annotated with goal hierarchies, sub-goal sequences, and temporal boundaries. This captures the causal and spatial reasoning chain that connects individual actions into coherent long-horizon behaviour - the kind of knowledge a humanoid robot must possess to operate autonomously in a shared public space. RoboVQA ~\cite{robovqa} is the closest comparator on the VLM fine-tuning axis, providing 829K auto-generated QA pairs for robot manipulation in
corporate office buildings; yet it is purely exocentric, single-domain, and lacks explicit reasoning chains.

The closest existing retail-specific embodied benchmark is Sari Sandbox / SariBench ~\cite{saribench}, a photorealistic Unity simulation of a Philippine convenience store with 100 VR human demonstrations across 250+ interactive products. While SariBench evaluates goal-directed agent task completion (navigation, object retrieval, product comparison), it is entirely simulation-based, egocentric-only, and an evaluation benchmark rather than a training corpus - providing no VLM instruction pairs, no temporal or physics probes, and no exocentric infrastructure views. It therefore differs substantially from PRISM in both modality and purpose.

In summary, prior datasets each capture a part of the problem, large-scale egocentric undestanding, ego-exo correspondence, exocentric retail action recognition or synthetic retail evaluation. PRISM ties these threads together in a single real-world training corpus. It is the first dataset to combine a domain-specific retail deployment scenario with multi-view panoramic video, ontology-structured supervision and SFT-ready instruction data for VLM adaptation.

\input{tables/dataset_comparison}

\nbf{Self-Supervised Visual Pretext Tasks}
Self-supervised pretext tasks have long been used to learn visual representations without manual annotation. Arrow-of-time prediction ~\cite{arrow_of_time} and jigsaw puzzle solving ~\cite{jigsaw} are classic examples that encourage models to learn temporal coherence and spatial structure, respectively. While these tasks are often used for representation learning, we instead repurpose them as instruction-tuning data. In PRISM, self-supervised transformations become question-answer pairs that can be incorporated directly into the SFT pipeline, allowing phyiscally grounded pretext signals to contribuie to embodied VLM adaptation.

%% file: tables/cosmos_comparison.tex
{\color{blue}
\begin{table}[t]
\centering
\caption{\textbf{PRISM vs.\ Cosmos-Reason1 training data by capability domain.}
Cosmos-Reason1 ~\cite{cosmos_reason1} SFT counts are taken from the
publicly released HuggingFace datasets.
\textcolor{black}{While AgiBot World includes retail scenes as one of five deployment domains ($*$), its SFT samples target general manipulation skills and are not structured as retail-domain perceptual QA; no dataset contributing to Cosmos-Reason1 provides retail-targeted VLM SFT.} PRISM contributes 270K retail-specific SFT pairs spanning all three domains. $\dagger$~Cosmos Physical Common Sense covers generic scenes; no retail or robot-deployment content is included. $\ddagger$~Novel tasks entirely absent from Cosmos-Reason1 training. \textcolor{black}{$*$~AgiBot World includes a retail deployment domain ($\sim$1 of 5 domains, supermarket shelf-picking and restocking tasks), but contributes no retail-targeted VLM SFT to Cosmos-Reason1 training.}}
\label{tab:cosmos_comparison}
\resizebox{\linewidth}{!}{%
\begin{tabular}{llrllr}
\toprule
\multirow{2}{*}{\textbf{Domain}} &
\multicolumn{2}{c}{\textbf{Cosmos-Reason1 (base model)}} & &
\multicolumn{2}{c}{\textbf{PRISM (ours)}} \\
\cmidrule{2-3}\cmidrule{5-6}
 & Source datasets & SFT samples & & Tasks & SFT samples \\
\midrule
Embodied  & BridgeV2, AgiBot$^*$,  & $\sim$1{,}720K & & ER-1--ER-9 (9 tasks)   & 93{,}757 \\
Reasoning & HoloAssist, RoboVQA   & \textcolor{black}{(general manip.; no retail SFT)}  & & incl.\ multi-actor$^\ddagger$, cross-view$^\ddagger$  &  \\
\midrule
Common    & Physical CS            & $\sim$1{,}800K$^\dagger$ & & CS-U-1/2, CS-R-1--4    & 47{,}778 \\
Sense +   & (curated video + MCQ) & (general scenes) & & SP-1, SP-2$^\ddagger$ (360\textdegree{}) & 11{,}862 \\
Spatial   &                        &                  & & incl.\ depth-grounded, exo spatial$^\ddagger$ & \\
\midrule
Intuitive & Arrow-of-Time,         & $\sim$51K      & & IP-1, IP-1 CoT, IP-2  & 91{,}000 \\
Physics   & Spatial Puzzles,       & (general)      & & IP-1 Exo+$^\ddagger$, CoT variants$^\ddagger$ & \\
          & Object Permanence      &                & &                        &          \\
\midrule
MCQ       & MCQ (general)          & $\sim$1{,}800K & & MCQ Overlay & 25{,}000 \\
\midrule
\textbf{Retail-targeted SFT} & \textit{(none)} & \textbf{0} & & \textbf{All 20 tasks} & \textbf{270K} \\
\bottomrule
\end{tabular}%
}
\vspace{-0.05in}
\end{table}
}

%% file: tables/dataset_comparison.tex
{\color{blue}
\begin{table}[t]
\centering
\caption{\textbf{Comparison with related embodied video datasets.}
PRISM is the only dataset that simultaneously covers retail deployment,
all three viewpoint modalities (ego, fixed exo, 360\textdegree{} exo),
explicit chain-of-thought supervision, and a VLM SFT-ready format.
\checkmark~=~yes; \texttimes~=~no.}
\label{tab:dataset_comparison}
\resizebox{\linewidth}{!}{%
\begin{tabular}{lrrcccc}
\toprule
\textbf{Dataset} & \textbf{Scale} & \textbf{\#Tasks} & \textbf{Viewpoint} & \textbf{Domain} & \textbf{CoT} & \textbf{SFT} \\
\midrule
Ego4D~\cite{ego4d}           & 3{,}670 h   & 5  & Ego               & General       & \texttimes & \texttimes \\
Ego-Exo4D~\cite{egoexo4d}    & 1{,}286 h   & 4  & Ego + Exo         & Skilled act.  & \texttimes & \texttimes \\
RoboVQA~\cite{robovqa}       & 829K pairs  & 6  & Exo               & Office robot  & \texttimes & \checkmark \\
SariBench~\cite{saribench}   & 100 demos   & 3  & Ego (sim.)        & Retail (sim.) & \texttimes & \texttimes \\
\midrule
\textbf{PRISM (ours)}        & \textbf{270K samples} & \textbf{20} & \textbf{Ego+Exo+360\textdegree{}} & \textbf{Retail} & \checkmark & \checkmark \\
\bottomrule
\end{tabular}%
}
\vspace{-0.05in}
\end{table}
}

%% file: 03_method.tex
\section{PRISM Dataset}
\label{sec:method}

\input{figs/examples_new_tasks}

\subsection{Data Collection Ontology}
\label{sec:data_collection}
PRISM is derived from video recorded across five structurally diverse retail environments spanning distinct store layouts, lighting conditions, aisle configurations, and product category distributions. This diversity is deliberate: by capturing variation across environments rather than within a single controlled deployment, PRISM produces training data that reflects the real distribution of structured service settings rather than a single site's idiosyncrasies. Data is captured from two complementary camera systems.


\nbf{Egocentric cameras} Wearable head-mounted cameras are used to capture the first-person perspective of human actor performing diverse shopping tasks in real-world retail grocery stores. These include entering and navigating the store, browsing shelves, approaching and selecting products, picking up and placing back items, carrying products, moving between aisles, and interacting with baskets, carts, shelves and checkout areas. The egocentric view captures the scene from the actor's first-person perspective, together with the natural hand-object interactions and local decision points that arise during shopping episodes. Egocentric clips are captured using a GoPro camera and during finetuning, they are encoded at 480p, 4~fps using H.264.

\nbf{Exocentric cameras} Exocentric views are captured using the proprietary ALIA 360\textdegree{}~\cite{DreamVuAlia} camera which provides multi-view and panoramic views of the scene through a multi-camera arrangement. Unlike the egocentric view, which follows the actor's local perspective, the exocentric view captures the broader scene, including aisle structure, shelf layout, navigation corridors and the movement of multiple people within the store. These views provide stable scene-level context for understanding spatial relationships, trajectories, multi-actor activity and environment-level behaviour during shopping episodes. During finetuning, the exocentric clips are encoded at 4-fps using H.264. 

Together, these two modalities provide complementary information about embodied activity in retail environments. Combining both views enables PRISM to support training and evaluation across fine-grained action understanding as well as scene-level spatial and behavioural reasoning. 

\subsection{Knowledge Ontology}
\label{sec:tasks}

PRISM is organized around a novel knowledge ontology for physical AI with three complementary and fundamental dimensions: spatial knowledge, temporal and physical knowledge, and embodied action knowledge. Spatial knowledge captures the understanding of 3D scene geometry, layout and relative structure. Temporal and physical knowledge captures reasoning about causality, motion, ordering, and physical constraints. Embodied action knowledge captures understanding of actions, goals, task progress, and activity across viewpoints. We draw inspiration from \cite{cosmos_reason1} for the knowledge ontology, and extend it to exocentric perspective and novel knowledge dimensions.

We operationalize our ontology through 20+ targeted capability probes spanning four domains - Embodied Reasoning (ER), Common Sense (CS), Spatial Perception (SP) and Intuitive Physics (IP) together with multiple-choice-question (MCQ) format training. These probes are expressed in three SFT formats: open-ended understanding, chain-of-thought reasoning with $\langle$think$\rangle$ tags, and MCQ answering. \cref{tab:dataset_overview} provides a quantitative summary and representative examples for all tasks are shown in \cref{fig:examples}. The capability probes are described in detail in the following section:

\subsubsection{Embodied Reasoning (ER)}
\label{sec:er}

\nbf{ER-1: Next Subtask Prediction}\\
Given a video clip showing the current activity with structured context  such as the goal, the sub-goal, the scene description, and the hand states, the model is expected to predict the next subtask in the shopping workflow. This task evaluates sequential reasoning and action anticipation both of which are central for embodied task execution. Labels are derived from temporal sub-goal ordering in episode metadata. The probe contains 25,000 samples.

\nbf{ER-2: Task Completion Verification}\\
This is a binary classification probe where the model is expected to determine whether a subtask is \textit{completed} or \textit{ongoing} based on video evidence. Completion labels are automatically inferred from sub-goal transition boundaries: if a clip occurs at the end of sub-goal and transitions to the next, it is labeled completed. This probe contains 3,000 samples with a 59.3\%/40.7\% class split. The samples are generated with 6 question templates and 8 answer templates to improve lexical diversity.

\nbf{ER-3: Goal-Conditioned Action Reasoning}\\
Given a clip together with structured context such as the goal hierarchy, environment and hand states, the model is expected to reason about why the current action is being performed in the context of the broader task goal. This evaluates causal understanding of embodied actions rather than action recognition alone which is critical for autonomous decision-making. This probe contains 8,000 samples.

\nbf{ER-4: Exo-to-Ego Activity Matching \textnormal{(exocentric)}}
Given a third-person clip showing an actor from an exocentric view, the model is expected to identify the activity being performed and match it to structured egocentric activity descriptions. Clips are extracted from synchronized Ego+Exo video pairs, isolating the exocentric perspective. This provbe explicitly connects first-person and third-person activity understanding and evaluates whether the same underlying behaviour can be recognized across viewpoints. This probe contains 7,000 samples.

\nbf{ER-5: Hand Interaction Recognition}\\
This probe is centered on model generated descriptions of hand states. The hand interactions in the video are labeled across 13 normalized categories: \textit{idle, holding, grasping, reaching, placing, using\_phone, pushing\_cart, carrying, pointing, scanning, arranging, picking\_up,} and \textit{pulling}. The model is expected to label the activity in the video to the appropriate hand state. This probe contains 15,000 samples extracted in an activity-balanced manner - 50\% both-hands-active, 30\% one-hand-active, 20\% both-hands-idle.

\nbf{ER-6: Atomic Action Recognition \textnormal{(exocentric)}}\\
This probe focuses on recognizing and describing short atomic actions from exocentric trajectory annotations. Using clips extracted from exocentric cameras, the model is expected to provides dense action-level recognition from the estimated human trajectories. This probe contains 20,000 samples.

\nbf{ER-7: Atomic Action Reasoning \textnormal{(exocentric, CoT)}}\\
This probe extends from ER-6 by moving from action recognition to action interpretation. Given the action description, the model is expected to reason \textit{why} a observed action is being performed.  This probe contains 8,000 samples with $\langle$think$\rangle$ chain-of-thought tags.

\nbf{ER-8: Multi-Actor Scene Understanding \textnormal{(exocentric)}}\\
This probe is based on multi-actor recognition. The model is expected to describe what multiple people are doing simultaneously in the store  from an exocentric viewpoint. This probe targets scene-level activity understanding in shared environments, where multiple actions may unfold in parallel. This probe contains 8,000 samples with at least two visible people over a minimum temporal window of $3s$.

\nbf{ER-9: Social Navigation Reasoning \textnormal{(exocentric, CoT)}}\\
This probe is a reasoning variant of ER-8 where the model is expected to analyze social and spatial dynamics between multiple actors - reasoning about how people move relative to each other, how they share space, and how individual behavior reflects social navigation constraints. Using the same multi-person clips, this probe contains 3,000 samples.

\begin{table}[t]
\centering
\caption{\textbf{PRISM dataset overview.} All 20+ capability probes organized by knowledge domain, viewpoint, SFT format, allocated sample counts, and annotation method. \textit{Und.}~=~open-ended understanding; \textit{CoT}~=~chain-of-thought reasoning with $\langle$think$\rangle$ tags; \textit{MCQ}~=~multiple-choice (A/B/C/D). Sample allocations follow sqrt-proportional scaling with no single task exceeding 9.5\% of the total mix.}
\label{tab:dataset_overview}
\resizebox{\textwidth}{!}{%
\begin{tabular}{llcccrl}
\toprule
\textbf{Task} & \textbf{Domain} & \textbf{View} & \textbf{Format} & \textbf{Allocated} & \textbf{\% of Total} & \textbf{Annotation Method} \\
\midrule
ER-1: Next Subtask Prediction & ER & Ego & Und.+MCQ & 25,000 & 9.2\% & Gemini Robotics ER 1.5 \\
ER-2: Task Completion Verification & ER & Ego & Und & 3,000 & 1.1\% & Gemini Robotics ER 1.5 \\
ER-3: Goal-Conditioned Action Reasoning & ER & Ego & CoT & 8,000 & 3.0\% & LLM (Gemini Flash) \\
ER-4: Exo-to-Ego Activity Matching & ER & Ego+Exo & Und. & 7,000 & 2.6\% & LLM (Gemini Flash) + Video \\
ER-5: Hand Interaction Recognition & ER & Ego & Und.+MCQ & 15,000 & 5.5\% & Gemini Robotics ER 1.5 \\
ER-6: Atomic Action Recognition & ER & Exo & Und. & 20,000 & 7.4\% & Gemini Robotics ER 1.5 \\
ER-7: Atomic Action Reasoning & ER & Exo & CoT & 8,000 & 3.0\% & LLM (Gemini Flash) \\
ER-8: Multi-Actor Scene Understanding & ER & Exo & Und. & 8,000 & 3.0\% & Gemini Robotics ER 1.5 \\
ER-9: Social Navigation Reasoning & ER & Exo & CoT & 3,000 & 1.1\% & LLM (Gemini Flash) \\
\midrule
CS-U-1: Scene Description VQA & CS & Ego & Und.+MCQ & 12,000 & 4.4\% & LLM (Gemini Flash) \\
CS-U-2: Environment VQA & CS & Exo & Und. & 8,000 & 3.0\% & LLM (Gemini Flash) \\
CS-R-1: Spatial Reasoning (LLM CoT) & CS & Ego & CoT+MCQ & 12,000 & 4.4\% & LLM + Depth \\
CS-R-2: Affordance Reasoning & CS & Ego & CoT & 8,000 & 3.0\% & Gemini Robotics ER 1.5 + Depth \\
CS-R-4: Spatial Reasoning & CS & Exo & CoT & 2,000 & 0.7\% & Gemini Robotics ER 1.5 + Depth \\
CS-R-3: Causality Reasoning & CS & Ego & CoT & 6,000 & 2.2\% & LLM (Gemini Flash) \\
\midrule
SP-1: Relative Depth Reasoning & SP & Ego & Und. & 10,000 & 3.7\% & Template + Depth \\
SP-2: 360\textdegree{} Spatial Layout & SP & Exo & CoT+MCQ & 2,000 & 0.7\% & LLM + Depth \\
\midrule
IP-1: Arrow-of-Time & IP & Ego+Exo & Und. & 30,000 & 12.9\% & Self-supervised \\
IP-1 CoT: Physics Reasoning & IP & Ego+Exo & CoT & 49,000 & 18.1\% & Gemini Robotics ER 1.5 + Video \\
IP-2: Object Permanence & IP & Ego & Und. & 8,000 & 3.0\% & Self-supervised \\
\midrule
MCQ Overlay (6 tasks) & Mixed & Mixed & MCQ & 12,000 & 4.4\% & LLM (Gemini Flash) \\
MCQ Standalone (5 categories) & Mixed & Text & MCQ & 14,000 & 5.2\% & LLM (Gemini Flash) \\
\midrule
\multicolumn{4}{l}{\textbf{Grand Total (20+ tasks)}} & \textbf{270K} & \textbf{100\%} & \\
\bottomrule
\end{tabular}%
}
\end{table}

\subsubsection{Common Sense (CS)}
\label{sec:cs}

\nbf{CS-U-1: Scene Description VQA}\\
This probe contains 12,000 samples and consists of factual visual question-answer pairs about the observed scene, spanning five balanced categories: \textit{person activity} (21.2\%), \textit{object identification} (20.5\%), \textit{environment description} (20.0\%), \textit{obstacles} (19.4\%), and \textit{scene description} (18.9\%). The model is expected to exhibit broad scene understanding grounded in visible retail context. The probe consists of QA pairs and to reduce data leakage, we limit generation to 3 QA pairs per episode.

\nbf{CS-U-2: Environment Understanding VQA \textnormal{(exocentric)}}\\
This probe is an exocentric counterpart to CS-U-1, using exocentric footage. Given an exocentric clip, the model is expected to answer questions about the visible objects, spatial layout, lighting, obstacles and scene structure. This probe contains 8,000 VQA samples. 

\nbf{CS-R-1: Spatial Reasoning with LLM Chain-of-Thought}\\
This probe is focused on reasoning about the structural aspects of the scene. The model is expected to answer spatial questions about depth relationships, obstacle proximity, and scene layout while producing explicit chain-of-thought reasoning. We extract depth maps and compute 8 per-frame statistics (mean depth, variance, near/far-field percentiles, gradient magnitude, left-right asymmetry, top-bottom asymmetry, foreground-background ratio). Questions in this probe are balanced across categories: \textit{relative distance}, \textit{obstacle awareness}, \textit{scene layout}, and \textit{depth profile}. This probe contains 12,000 samples.

\nbf{CS-R-2: Affordance Reasoning \textnormal{(CoT)}}\\
This probe expects the model to reason about object affordances - what actions are possible on visible objects and surfaces, and under what constraints?. Given an egocentric clip together with depth context, the model is expected to reason about reachable objects, possible interactions such as grasping, pushing, pulling, or opening, and the spatial limitations governing those actions. The probe targets embodied common sense that is closely tied to action feasibility in the physical world. This probe contains 8,000 samples.

\nbf{CS-R-3: Causality Reasoning \textnormal{(CoT)}}\\
Given an egocentric clip with episode annotations, the model is expected to explain causal relationships between observed actions and their context - reasoning about why the person performed the observed action, what caused it, and what effects it is likely to procude. This probe focuses on everyday causal understanding in embodied retail scenarios and consists of 6,000 samples.

\nbf{CS-R-4: Spatial Reasoning \textnormal{(exocentric, CoT)}}\\
This probe extends spatial reasoning to exocentric viewpoints. The model is expected to analyze clips to reason about 3D scene layout, person positions, and depth relationships from a scene-centered perspective. This probe contains 2,000 samples.

\subsubsection{Spatial Perception (SP)}
\label{sec:sp}

\nbf{SP-1: Relative Depth Reasoning}\\
This probe is centered on relative depth ordering, depth gradients, foreground/background separation, and depth range characterization from egocentric video. The model is expected to reason spatial statistics across five balanced categories: region comparison, depth gradient, foreground/background analysis, nearest/farthest identification, and depth range characterization. This probe includes 10,000 samples.

\nbf{SP-2: 360\textdegree{} Spatial Layout Reasoning \textnormal{(exocentric)}}
This probe is applied on panoramic viewpoints. Alia's 360$^\circ$ omnidirectional views are used and the model is expected to produce spatial layout understanding with chain-of-thought. For each clip, depth maps are used to extract the same 8 depth statistics used in CS-R-1 to ground the spatial reasoning. The panoramic 360$^\circ$ perspective introduces spatial reasoning challenges that differ substantially from the forward-facing egocentric view. This probe contains approximately 2,000 samples.

\subsubsection{Intuitive Physics (IP)}
\label{sec:ip}

\nbf{IP-1: Arrow-of-Time (Egocentric and Exocentric)}\\
This is a binary classification probe where the model determines whether a video plays \textit{forward} or \textit{backward}. It is fully self-supervised as reversed clips are produced via temporal reversal, and labels are assigned by construction. Answers are shortened to direction-only (``The video is playing forward/backward.'') to prevent answer-style overlap with ER/CS scene descriptions. This probe  contains 30,000 samples with a 50/50 class balance.

\nbf{IP-1 Chain-of-Thought: Physics Reasoning (Egocentric, CoT)}\\
This probe is focused on temporal reasoning. For each IP-1 clip, the model is expected to generate chain-of-thought reasoning over three physics axes: gravity and momentum cues, body mechanics/biomechanics, and temporal flow and causality. These chains are intended to capture physically grounded reasoning that is distinct from ER and CS style scene understanding, and to encourage explanations based on the internal consistency of observed motion. This probe evaluates physically grounded interpretation from first-person motion cues as well as scene-level exocentric views. This probe contains 49,000 samples.

\nbf{IP-2: Object Permanence}\\
This is a self-supervised probe where the model is expected to affirm whether objects continue to exist even when temporarily out of frame. Every ER-1 egocentric clip is paired with a question about object permanence together with a canned answer asserting physical persistence. This probe introduces a simple but important aspect of intuitive physical understanding in everyday environments and contains 8,000 samples.

\nbf{MCQ Overlay}\\
This probe converts six existing open-ended tasks (ER-1, ER-5, CS-U-1, CS-R-1, SP-1, SP-2) into a multiple-choice format by pairing each video clip with four answer options one correct and three hard distractors generated by Gemini 2.5 Flash. The model receives the video together with the question and four labeled choices (A/B/C/D), and is expected to output a single letter. This format trains the model to perform constrained selection under visual grounding, complementing the open-ended generation format used elsewhere in PRISM. This probe contains 12,000 samples balanced across the six source tasks.

\nbf{MCQ Standalone}\\
This probe consists of text-only multiple-choice questions spanning five knowledge categories: retail operations, spatial reasoning, physics understanding, visual question answering, and action recognition. Unlike MCQ Overlay, these questions do not include video input, they test the model's retained domain knowledge and reasoning ability in a purely textual setting. Each question presents four options (A/B/C/D) with a single correct answer. The questions are designed to complement the video-grounded tasks by reinforcing conceptual understanding. This probe contains 14,000 samples.

\begin{table}[t]
\centering
\small
\caption{\textbf{PRISM scale statistics.} Token counts at 4\,fps with Qwen3VL tokenization.}
\label{tab:dataset_scale}
\begin{tabular}{lr}
\toprule
\textbf{Metric} & \textbf{Value} \\
\midrule
Total samples & 270K \\
Tasks (+ 2 eval-only) & 20+ \\
Video frames (4\,fps) & $\sim$11.8M \\
Total tokens & $\sim$730M (703M visual + 27M text) \\
\midrule
Open-ended / CoT / MCQ & 58.6\% / 30.4\% / 9.2\% \\
Domain: ER / CS / SP / IP / MCQ & 34.7 / 18.3 / 4.4 / 33.5 / 9.2\% \\
\bottomrule
\end{tabular}
\end{table}

\subsection{Annotation Pipeline}
\label{sec:annotation}

PRISM is constructed using five complementary annotation strategies that together provide broad coverage, scalable supervision and diverse training signals across the dataset.

\nbf{Metadata Extraction}
For ER-1, ER-2, ER-4, ER-5, ER-6, and ER-8, annotations are derived from structured metadata produced by Gemini Robotics ER 1.5 ~\cite{gemini_robotics} - a specialized model for embodied physical reasoning. These metadata include episode-level summaries that contain  goal hierarchies, sub-goal sequences, temporal boundaries, hand-state labels, exocentric trajectory annotations, and activity summary JSONs. We use template pools to increase linguistic diversity, for example through multiple question-answer template combinations for tasks such as ER-2 and ER-4. 

\nbf{LLM-Based Generation (Gemini 2.5 Flash)}
For CS-U-1, CS-U-2, CS-R-1, CS-R-3, SP-2, ER-3, ER-7, ER-9, ER-4 (expanded), and MCQ generation, we use Gemini 2.5 Flash ~\cite{gemini25} to generate QA pairs and chain-of-thought reasoning.  Prompts operate primarily over structured text annotations, with video conditioned prompting prompting used where the task requires direct visual grounding. MCQ overlay samples are generated by prompting Gemini to produce three hard distractors for every existing answer, and randomly assigning the correct option across A/B/C/D.. MCQ standalone samples are generated as text-only questions spanning five knowledge categories - retail, spatial, physics, visual QA and action recognition.

\nbf{Physics-Grounded Video Reasoning (Gemini Robotics ER 1.5)}
For CS-R-2, CS-R-4 and IP-1 CoT we use again utilize Gemini Robotics ER 1.5 . Unlike text-only LLM generation, these tasks require the model to process the actual video directly. Clips are uploaded through Gemini's video processing API, and the model generates physics-grounded reasoning about gravity, momentum, body mechanics, temporal causality, affordances and spatial layout. This produces richer chain-of-thought supervision than text-conditioned prompting.

\nbf{Depth-Aware Analysis}
Spatial reasoning tasks - CS-R-1, CS-R-2, CS-R-4, SP-1, SP-2 - are grounded in depth estimated from DepthCrafter ~\cite{depthcrafter}. For each clip, we compute spatial statistics such as mean depth, min/max, standard deviation, left-right and top-bottom quadrant means, center depth, and foreground-background separation.

\nbf{Self-Supervised Generation}
IP-1 and IP-2 do not require any annotation or API calls. Arrow-of-time labels are determined by construction (original vs.\ reversed), object permanence answers assert physical persistence by construction. All samples are generated through deterministic video/image transformations. IP-1 answers are deliberately shortened to direction-only - ``The video is playing forward/backward'' - to prevent overlap with ER/CS scene description answer styles.

\subsection{Data Format}
\label{sec:format}

All samples follow the Cosmos-Reason2 format ~\cite{cosmos_reason2} for supervised fine-tuning: a three-message conversation consisting of system, user and assistant turns. User messages contain interleaved video or image content together with task text. 

We use three response formats: ~\textit{Understanding} tasks produce direct open-ended answers, ~\textit{Reasoning} tasks  wrap the response in $\langle$\texttt{think}$\rangle$\ldots$\langle$/\texttt{think}$\rangle$ tags containing explicit chain-of-thought followed by the final answer, ~\textit{MCQ} tasks present four choices (A/B/C/D) in the question and expect a single-letter response. 

Following LLaVA-Video ~\cite{llava_video}, we target a balanced supervision mix of approximately $\sim$60\% open-ended, $\sim$30\% CoT, and $\sim$10\% MCQ for balanced format training. Videos are sampled at 4~fps (integer). A first-token diversity audit ensures no task has $>$50\% of answers sharing the same opening word, mitigating first-token bias during training.

%% file: figs/examples_new_tasks.tex

\begin{figure}[t]
\centering
\small
\setlength{\tabcolsep}{4pt}

\begin{tabular}{@{}p{0.22\textwidth} p{0.70\textwidth}@{}}
\toprule
\textbf{Frame} & \textbf{Task / Question / Answer} \\
\midrule

\vspace{0pt}\includegraphics[width=\linewidth]{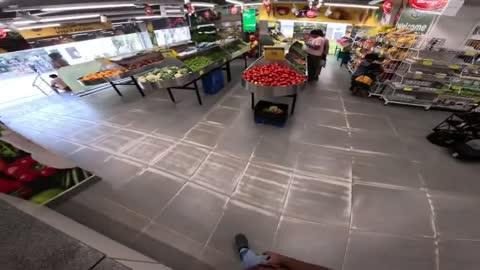} &
\vspace{0pt}\textbf{ER-1: Next Subtask Prediction} \textit{(Ego)} \newline
\textbf{Q:} What is the next subtask? \newline
\textbf{A:} Navigate to Dry Goods aisle. Walk forward into an aisle.
\\[2pt] \hline \\[-8pt]

\vspace{0pt}\includegraphics[width=\linewidth]{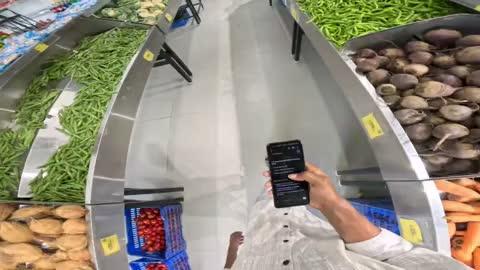} &
\vspace{0pt}\textbf{ER-2: Task Completion Verification} \textit{(Ego)} \newline
\textbf{Q:} Has ``Navigate Aisle while Reviewing List'' been completed? \newline
\textbf{A:} No, the person walks from produce towards freezers.
\\[2pt] \hline \\[-8pt]

\vspace{0pt}\includegraphics[width=\linewidth]{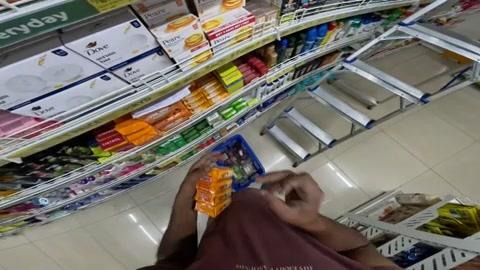} &
\vspace{0pt}\textbf{ER-3: Goal-Conditioned Action Reasoning} \textit{(Ego, CoT)} \newline
\textbf{Q:} Based on the person's current task, where should they navigate to next in the store? \newline
\textbf{A:} \textit{$\langle$think$\rangle$The person has just finished inspecting the Heritage Platinum Basmati Rice. Their current sub-goal is ``Examine Product Details,'' which they have completed. The next logical step is to find another product to compare or move on to the next item on their list.$\langle$/think$\rangle$} Continue down the current aisle to explore other rice options or adjacent product categories.
\\[2pt] \hline \\[-8pt]

\vspace{0pt}\includegraphics[width=\linewidth]{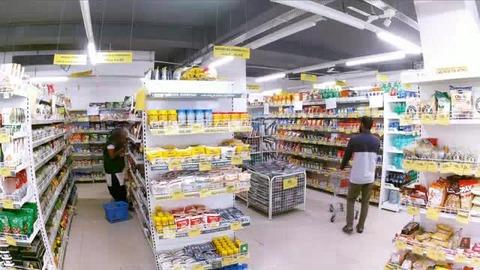} &
\vspace{0pt}\textbf{ER-4: Exo-to-Ego Activity Matching} \textit{(Exo)} \newline
\textbf{Q:} What activity is the person performing? \newline
\textbf{A:} The worker is scanning shelf labels and restocking items in the beverage aisle.
\\[2pt] \hline \\[-8pt]

\vspace{0pt}\includegraphics[width=\linewidth]{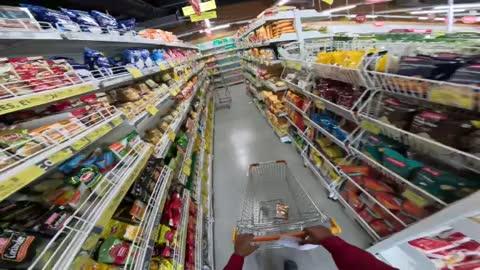} &
\vspace{0pt}\textbf{ER-5: Hand Interaction Recognition} \textit{(Ego)} \newline
\textbf{Q:} Describe the hand interactions visible in this video. \newline
\textbf{A:} Both hands holding an item. Pushes cart down the aisle.
\\[2pt] \hline \\[-8pt]

\vspace{0pt}\includegraphics[width=\linewidth]{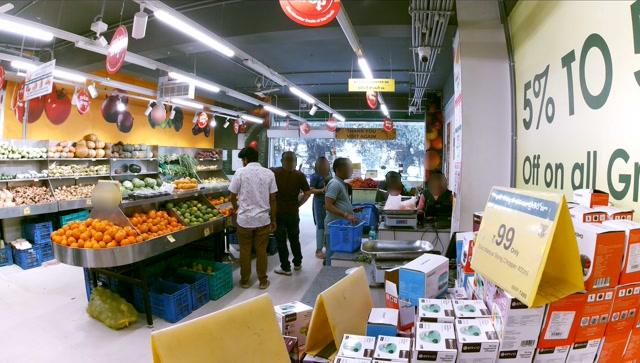} &
\vspace{0pt}\textbf{ER-6: Atomic Action Recognition} \textit{(Exo)} \newline
\textbf{Q:} Based on the video, what action is the person currently performing? \newline
\textbf{A:} The action being performed by the man is: reaching toward basket (target: blue shopping basket). Extends right arm towards the blue basket.
\\[2pt] \hline \\[-8pt]

\vspace{0pt}\includegraphics[width=\linewidth]{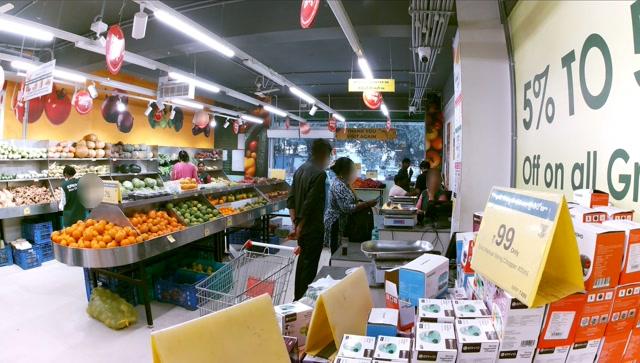} &
\vspace{0pt}\textbf{ER-7: Atomic Action Reasoning} \textit{(Exo, CoT)} \newline
\textbf{Q:} What is the likely reason this person is doing what they're doing in this scene? \newline
\textbf{A:} \textit{$\langle$think$\rangle$The staff member is wearing a green vest, which is often an indicator of employee attire in a retail setting. They are bending towards the produce shelf. The action of bending with knees and hips flexing while maintaining a straight back is a biomechanically efficient way to lower one's body to inspect or arrange items on lower shelves.$\langle$/think$\rangle$} The staff member is restocking or inspecting produce items on the lower shelf as part of routine store maintenance.
\\
\bottomrule
\end{tabular}

\caption{\textbf{PRISM capability probe examples (Part~1 of 3).} Each row shows a representative video frame alongside the full question and model answer. CoT tasks include the chain-of-thought in $\langle$think$\rangle$ tags (shown in italics) before the final answer.}
\label{fig:examples}
\end{figure}

\begin{figure}[t]
\ContinuedFloat
\centering
\small
\setlength{\tabcolsep}{4pt}

\begin{tabular}{@{}p{0.22\textwidth} p{0.70\textwidth}@{}}
\toprule
\textbf{Frame} & \textbf{Task / Question / Answer} \\
\midrule

\vspace{0pt}\includegraphics[width=\linewidth]{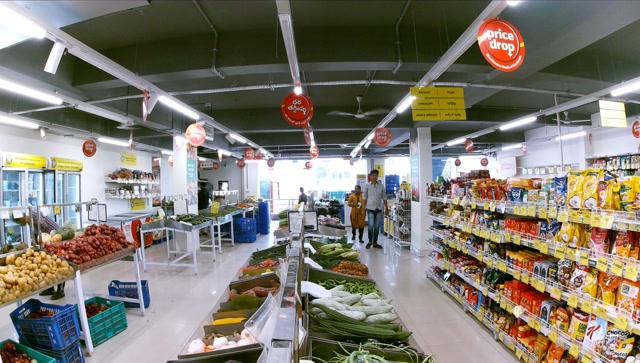} &
\vspace{0pt}\textbf{ER-8: Multi-Actor Understanding} \textit{(Exo)} \newline
\textbf{Q:} List all the people visible and their current activities in this retail scene. \newline
\textbf{A:} The footage captures 2 people in the retail environment. Person~1: A female store employee walks along the produce display, stopping to arrange and organize vegetables on the shelves and bins. Person~2: A man walks forward in the aisle, carrying a blue plastic shopping basket.
\\[2pt] \hline \\[-8pt]

\vspace{0pt}\includegraphics[width=\linewidth]{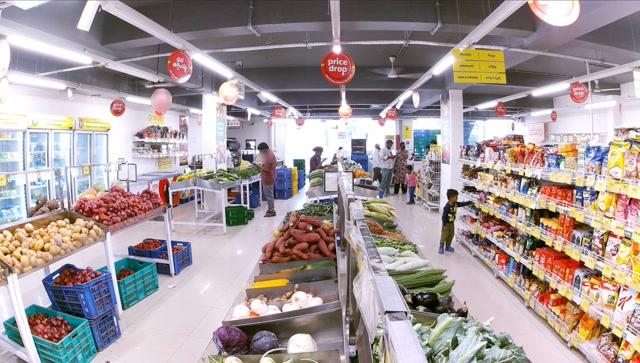} &
\vspace{0pt}\textbf{ER-9: Social Navigation Reasoning} \textit{(Exo, CoT)} \newline
\textbf{Q:} Describe the actions of each individual visible in this store camera view. \newline
\textbf{A:} \textit{$\langle$think$\rangle$From this camera view, 4 people can be identified. Person~1 walks into the produce section and interacts with staff. Person~2 is a staff member assisting at the vegetable display. Person~3 browses independently near the refrigerated section. Person~4 navigates toward the checkout area.$\langle$/think$\rangle$} Four individuals occupy the store with distinct goals: one customer selects produce with staff assistance, another browses independently, and a fourth heads toward checkout.
\\[2pt] \hline \\[-8pt]

\vspace{0pt}\includegraphics[width=\linewidth]{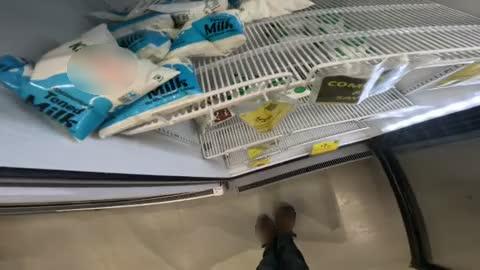} &
\vspace{0pt}\textbf{CS-U-1: Scene Description VQA} \textit{(Ego)} \newline
\textbf{Q:} Describe the scene visible in this video. \newline
\textbf{A:} Person at refrigerator door, looking at milk and curd.
\\[2pt] \hline \\[-8pt]

\vspace{0pt}\includegraphics[width=\linewidth]{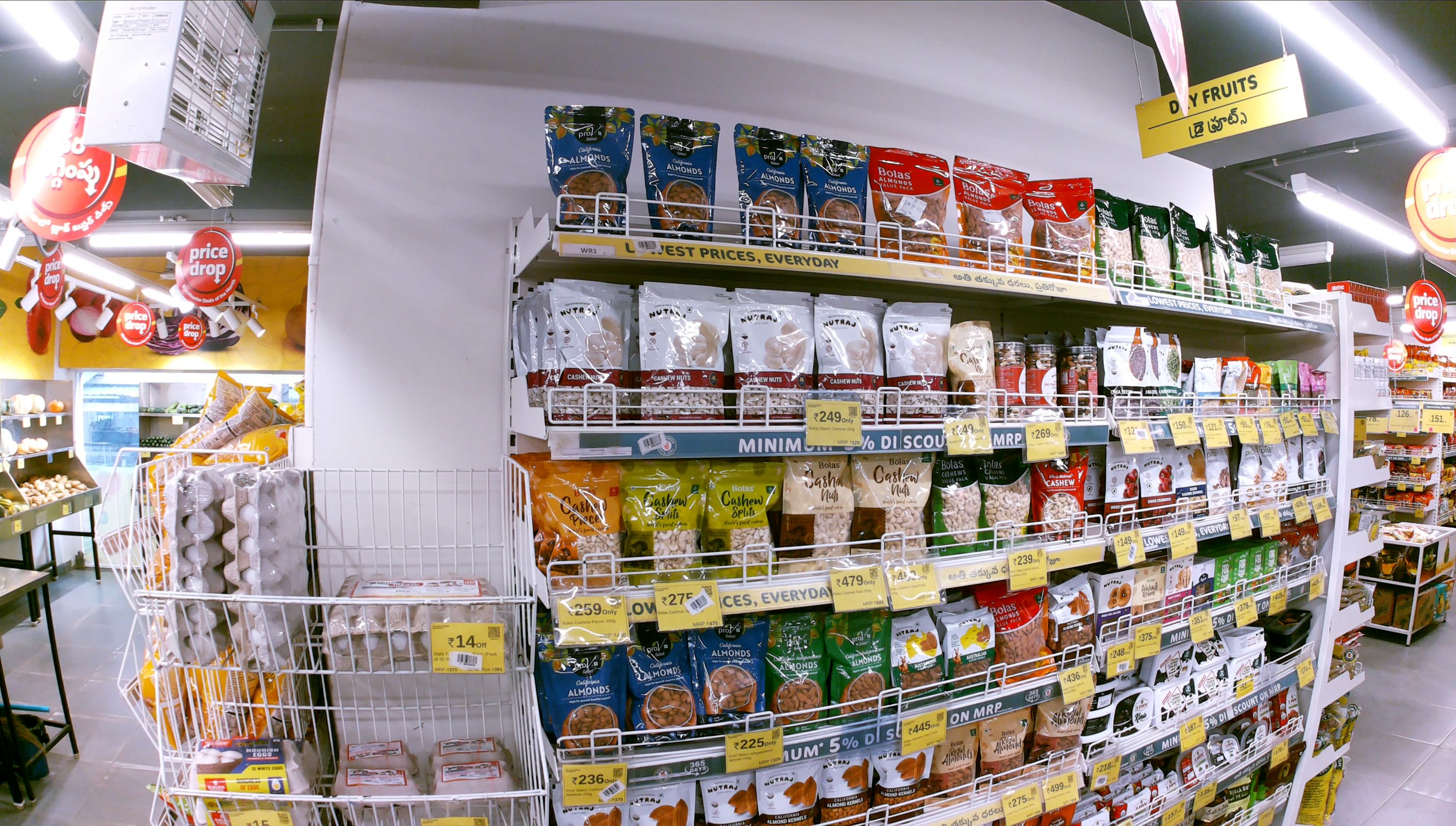} &
\vspace{0pt}\textbf{CS-U-2: Environment Understanding VQA} \textit{(Exo)} \newline
\textbf{Q:} What is the typical arrangement of products on the shelving units in this grocery store aisle? \newline
\textbf{A:} The shelving units display a variety of packaged food items, primarily snacks, arranged by category. Products are neatly stocked and often stacked to maximize shelf space, with labels facing forward for easy identification by customers.
\\[2pt] \hline \\[-8pt]

\vspace{0pt}\includegraphics[width=\linewidth]{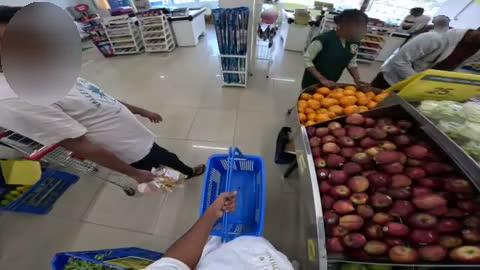} &
\vspace{0pt}\textbf{CS-R-1: Spatial Reasoning} \textit{(Ego, CoT)} \newline
\textbf{Q:} Are items close or spread out in this scene? \newline
\textbf{A:} \textit{$\langle$think$\rangle$Mean depth 0.17, most objects within near field...$\langle$/think$\rangle$} Close range. Items are within arm's reach.
\\[2pt] \hline \\[-8pt]

\vspace{0pt}\includegraphics[width=\linewidth]{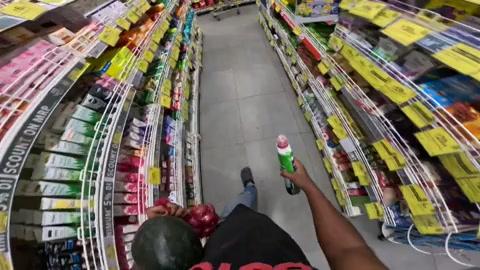} &
\vspace{0pt}\textbf{CS-R-2: Affordance Reasoning} \textit{(Ego, CoT)} \newline
\textbf{Q:} Based on the spatial layout, what affordances are available to the person? \newline
\textbf{A:} \textit{$\langle$think$\rangle$The shelves on both sides are within immediate reaching distance, affording grasping of products like snacks, beverages, and household items. The floor affords walking through the aisle. The shopping basket provides containment for selected items.$\langle$/think$\rangle$} The environment affords navigation through aisles and grasping items from shelves and bins. The basket facilitates collection of products.
\\[2pt] \hline \\[-8pt]

\vspace{0pt}\includegraphics[width=\linewidth]{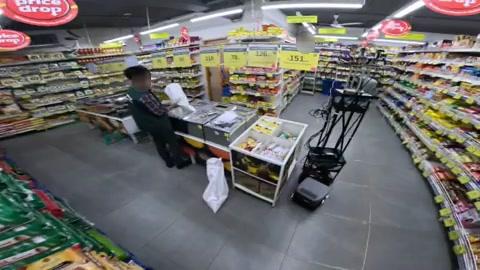} &
\vspace{0pt}\textbf{CS-R-3: Causality Reasoning} \textit{(Ego, CoT)} \newline
\textbf{Q:} Analyze the cause-and-effect: why did this action happen? \newline
\textbf{A:} \textit{$\langle$think$\rangle$The person is in an aisle stocked with dry goods and beverages. Their goal is to perform grocery shopping, which necessitates navigating aisles to find items on their list. Walking down this aisle directly progresses through the store's layout toward the target items.$\langle$/think$\rangle$} To find items on their shopping list located in the dry goods and beverages aisle.
\\
\bottomrule
\end{tabular}

\caption[]{\textbf{PRISM capability probe examples (Part~2 of 3).} Continued from above.}
\end{figure}

\begin{figure}[t]
\ContinuedFloat
\centering
\small
\setlength{\tabcolsep}{4pt}

\begin{tabular}{@{}p{0.22\textwidth} p{0.70\textwidth}@{}}
\toprule
\textbf{Frame} & \textbf{Task / Question / Answer} \\
\midrule

\vspace{0pt}\includegraphics[width=\linewidth]{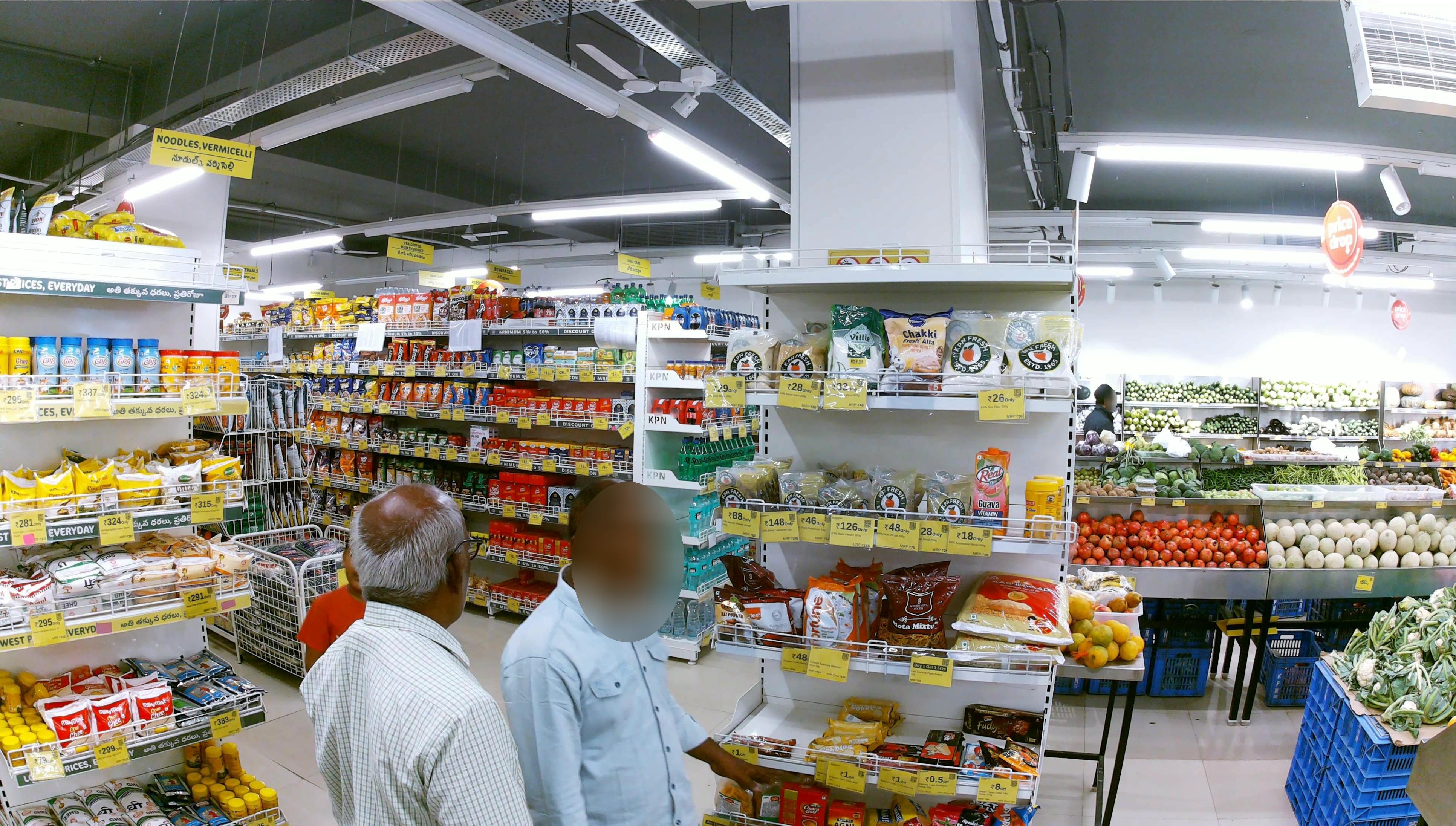} &
\vspace{0pt}\textbf{CS-R-4: Spatial Reasoning} \textit{(Exo, CoT)} \newline
\textbf{Q:} Describe the typical person-object interaction required to select items from the bulk bins and refrigerated cases. \newline
\textbf{A:} \textit{$\langle$think$\rangle$To interact with the bulk bins, a customer would reach over the counter, lift the transparent lid, and use a scoop to retrieve the product. For refrigerated cases, the customer would open one of the glass doors to access and select items.$\langle$/think$\rangle$} Customers interact with bulk bins by lifting lids and scooping products, and with refrigerated cases by opening glass doors to retrieve items.
\\[2pt] \hline \\[-8pt]

\vspace{0pt}\includegraphics[width=\linewidth]{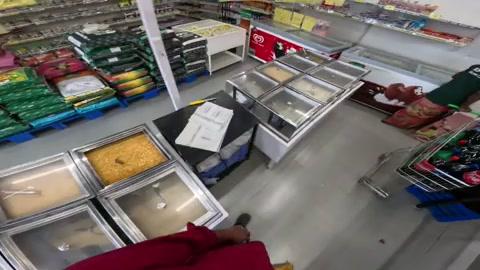} &
\vspace{0pt}\textbf{SP-1: Relative Depth Reasoning} \textit{(Ego)} \newline
\textbf{Q:} Does this scene have a wide or narrow depth range? \newline
\textbf{A:} This scene has a moderate depth range from 0.05 to 0.78. There is a mix of closer and farther objects, but the distances do not span an extreme range.
\\[2pt] \hline \\[-8pt]

\vspace{0pt}\includegraphics[width=\linewidth]{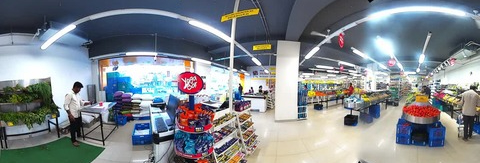} &
\vspace{0pt}\textbf{SP-2: 360\textdegree{} Spatial Layout Reasoning} \textit{(Exo, CoT)} \newline
\textbf{Q:} Describe the spatial layout and obstacle proximity. \newline
\textbf{A:} \textit{$\langle$think$\rangle$Mean depth 0.43, bimodality 0.07...$\langle$/think$\rangle$} Open central aisle, shelves on both sides at moderate distance.
\\[2pt] \hline \\[-8pt]

\vspace{0pt}\includegraphics[width=\linewidth]{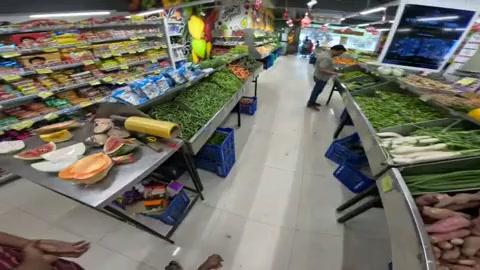} &
\vspace{0pt}\textbf{IP-1: Arrow-of-Time (Physics Reasoning)} \textit{(Ego, CoT)} \newline
\textbf{Q:} Watch this video and determine if it is playing forward or backward. Analyze the body mechanics and biomechanical cues. \newline
\textbf{A:} \textit{$\langle$think$\rangle$While the walking motion appears smooth, the foot placement and overall gait reveal subtle inconsistencies. The natural push-off and weight transfer of a forward stride are absent, giving the movement a slightly unnatural, gliding quality.$\langle$/think$\rangle$} The video is playing backward.
\\[2pt] \hline \\[-8pt]

\vspace{0pt}\includegraphics[width=\linewidth]{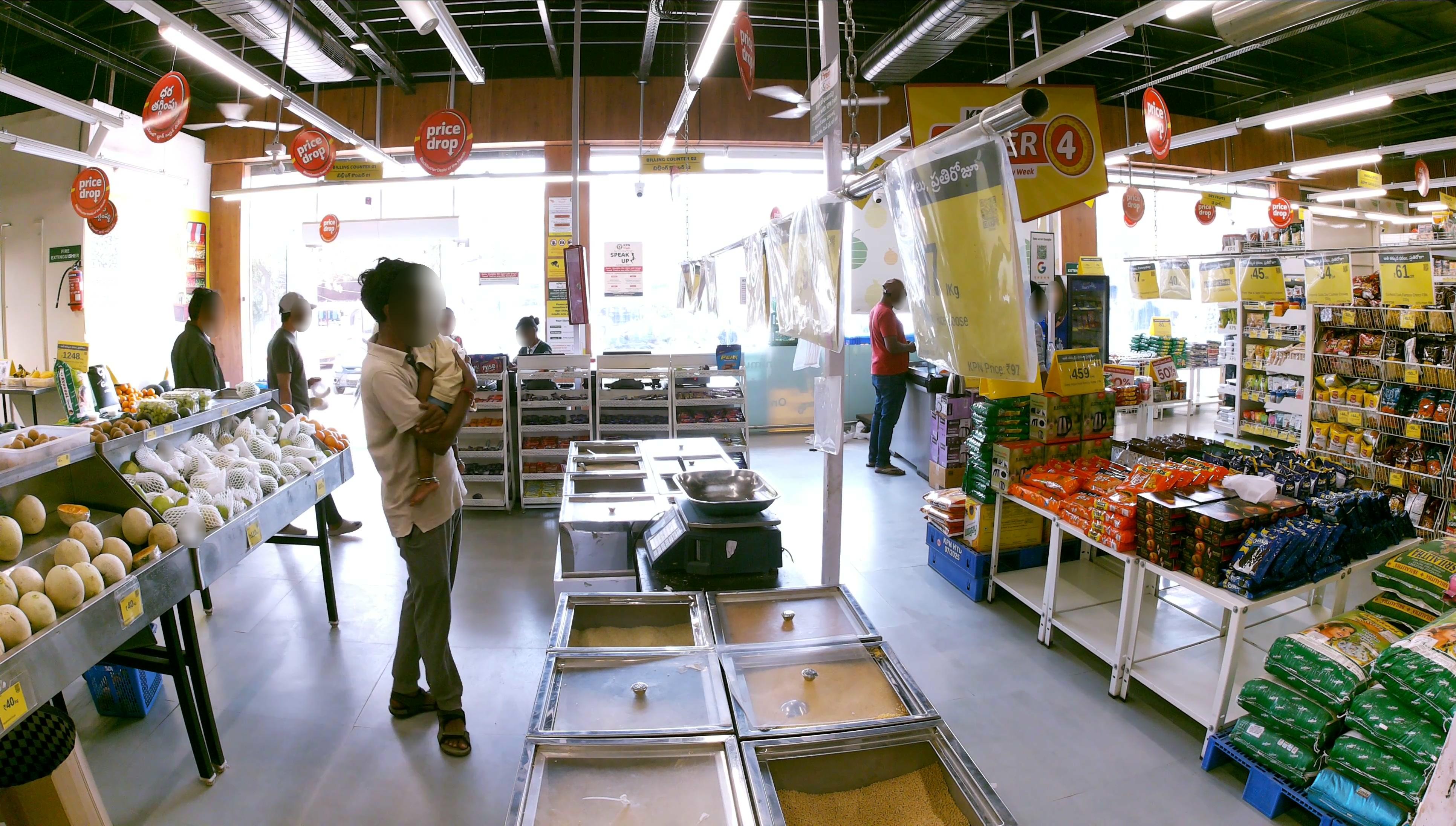} &
\vspace{0pt}\textbf{IP-1: Arrow-of-Time (Physics Reasoning)} \textit{(Exo, CoT)} \newline
\textbf{Q:} Watch this video and determine if it is playing forward or backward. Analyze the temporal flow and causal relationships. \newline
\textbf{A:} \textit{$\langle$think$\rangle$A person enters the frame from the left and walks toward the right. The legs move in a natural forward progression. There are no visual cues of reverse motion or unnatural movement patterns.$\langle$/think$\rangle$} The video is playing forward.
\\[2pt] \hline \\[-8pt]

\vspace{0pt}\includegraphics[width=\linewidth]{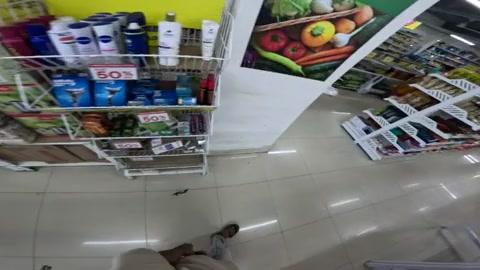} &
\vspace{0pt}\textbf{IP-2: Object Permanence} \textit{(Ego)} \newline
\textbf{Q:} Something left the field of view. Does the object continue to exist when not visible? \newline
\textbf{A:} The object still exists. In a retail environment, items on shelves, carts, and people maintain their physical presence even when they leave the camera frame. Object permanence is a basic physical principle.
\\[2pt] \hline \\[-8pt]

\vspace{0pt}\includegraphics[width=\linewidth]{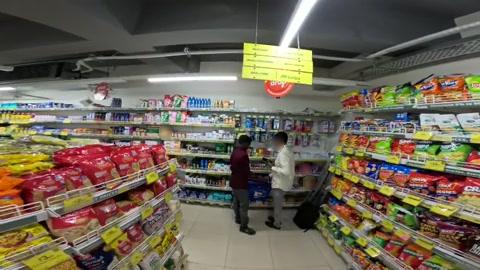} &
\vspace{0pt}\textbf{MCQ Overlay: Multiple-Choice with Video} \textit{(Mixed)} \newline
\textbf{Q:} What hand activity can you observe in this video? \newline
A.~One hand is actively engaged in gesturing while the other is still. \newline
B.~Both hands are idle and resting at the person's sides. \newline
C.~The person is holding an object in their right hand and gesturing with the left. \newline
D.~The person is using both hands to clap rhythmically. \newline
\textbf{A:} B
\\
\bottomrule
\end{tabular}

\caption[]{\textbf{PRISM capability probe examples (Part~3 of 3).} CS-R-4 and SP tasks test spatial understanding. IP-1 CoT variants demonstrate physics-grounded reasoning about temporal direction. IP-2 evaluates object permanence. MCQ Overlay converts open-ended tasks into four-choice format.}
\end{figure}

%% file: 04_experiments.tex
\section{Experimental Setup and Results}
\label{sec:experiments}

\subsection{Base Model and Fine-Tuning}
\label{sec:finetuning}

We use Cosmos-Reason2-2B~\cite{cosmos_reason2} as our base model, a vision-language model developed by NVIDIA for physical AI applications and built on the Qwen3-VL architecture~\cite{bai2025qwen3}. Cosmos-Reason2-2B is a suitable base model for PRISM because it is explicitly designed for video language reasoning in embodied settings, rather than for open-domain visual question answering alone. Its pretraining strategies and instruction-tuning methods emphasize physical scene understanding, action interpretation, and temporally grounded reasoning, making it a strong foundation for studying domain-specific SFT. The model processes interleaved video frames and text through a shared transformer backbone with 1.53B base parameters.

We fine-tune Cosmos-Reason2-2B using parameter efficient adaptation, allowing us to isolate the contribution of PRISM training mixtures while keeping the training setup practical and consistent across experiments. Our primary configuration uses BF16 LoRA without quantization with rank 32 and alpha 64 applied to all language model linear layers including the query, key, value, output, gate, up an down projections. This results in 49.3M trainable parameters corresponding to 1.98\% of the full model. Training uses AdamW ~\cite{adamw} with learning rate $1 \times 10^{-4}$, cosine schedule, 5\% warmup, batch size 1 with gradient accumulation 8 across 4 GPUs yielding an effective batch-size of 32, and train in bfloat16 precision throughout. All models are trained for one epoch on each data mixture with gradient checkpointing enabled. 

The primary experiments are conducted on 4$\times$ NVIDIA RTX PRO 6000 Blackwell GPUs (96\,GB each) using distributed data parallel training, achieving approximately 14\,s per step. To compare against a lower-memory alternative, we additionally train a QLoRA variant on 2$\times$ NVIDIA L40S 48\,GB GPUs. This allows us to examine whether the main trends we observe are robust to a more memory-constrained fine-tuning setup.


\subsection{Evaluation Protocol}
\label{sec:evaluation}

We evaluate all models on held-out validation sets across all 20+ capabilities (250 samples per task) using \textbf{MCQ accuracy}. For each open-ended task, we generate multiple-choice validation sets (4 options, 1 correct) via Gemini 2.5 Flash. Classification tasks (ER-2, IP-1 Ego, IP-1 Exo) use exact-match accuracy.

For chain-of-thought (CoT) tasks, we additionally employ GPT-4o as an automated judge to score the full model output including the intermediate reasoning within $\langle$think$\rangle$ tags and the final answer---on a normalized 0--100 scale, providing a complementary measure of open-ended generation quality beyond MCQ accuracy.

We report per-capability scores, per-domain scores (macro-average over capabilities in each domain), and overall score (macro-average over all capabilities). The zero-shot Cosmos-Reason2-2B model (\textbf{Baseline}) provides the lower bound.

%% file: 05_results.tex


\subsection{Results}
\label{sec:main_results}

The MCQ accuracies for each task evaluated for both zero-shot baseline  and the PRISM-fine-tuned model are given in \cref{tab:main_results}. Fine-tuning on PRISM yields substantial improvements across all capability domains, with an average gain of \textbf{+23.8\%} over the pre-trained baseline (from \textbf{62.8\%} to \textbf{86.6\%}). As different tasks start with highly varying baseline accuracy, a better way to understand the impact of the fine-tuning is by looking at the reduction in error rate. For the harder domains such as embodied reasoning, the error rate goes down from \textbf{45.5\%} to \textbf{9.1\%}, or by a \textbf{factor of five}. The average reduction in error rate across all tasks is \textbf{66.6\%}. Table~\ref{tab:main_results} also presents the average accuracies and error ratio for each domain as well. Let us now take a closer look at each domain separately.

\begin{table*}[htp]
\centering 
\caption{\textbf{Performance Comparison between Baseline} (Zero-Shot Cosmos-Reason2-2B) and \textbf{PRISM} (Our model fine-tuned on the full 270K training corpus) \textbf{on all evaluation tasks.} Best score per task in \textbf{bold}. The Error Ratio shows the ratio of PRISM fine-tuned error rate to the baseline error rate.}
\label{tab:main_results}
\begin{tabular}{lccccc}
\toprule
\textbf{Domain-TaskID: Task} & \textbf{View} & \textbf{Baseline (\%)} & \textbf{PRISM (\%)} & \textbf{$\Delta$(\%)} & \textbf{Error Ratio} \\
\midrule
ER-1: Next Subtask Prediction       & Ego  & 30.8 & \textbf{76.8} & +46.0 & $0.34$\\
ER-2: Task Completion                & Ego  & 53.6 & \textbf{87.6} & +34.0  & $0.27$\\
ER-3: Action Reasoning               & Ego  & 32.4 & \textbf{95.2} & +62.8  & $0.07$\\
ER-4: Cross-View Matching$^\star$    & Ego+Exo  & 42.9 & \textbf{91.0} & +48.1  & $0.16$\\
ER-5: Hand Interaction               & Ego  & 61.6 & \textbf{95.6} & +34.0  & $0.11$\\
ER-6: Atomic Action$^\star$          & Exo  & 54.0 & \textbf{84.8} & +30.8 & $0.33$\\
ER-7: Atomic Action CoT$^\star$      & Exo  & 41.6 & \textbf{96.8} & +55.2 & $0.05$\\
ER-8: Multi-Actor$^\star$            & Exo  & 85.2 & \textbf{94.0} & +8.8 & $0.41$\\
ER-9: Social Reasoning$^\star$       & Exo  & 88.8 & \textbf{96.4} & +7.6 & $0.32$\\
\cmidrule{4-6}
\textit{ER Average (9 tasks)} & & \textit{54.5} & \textit{\textbf{90.9}} & \textit{+36.4} & $0.20$\\
\midrule
CS-U-1: Scene VQA                    & Ego  & 66.8 & \textbf{88.4} & +21.6 & $0.35$\\
CS-U-2: Environment VQA$^\star$      & Exo  & 95.6 & \textbf{99.6} & +4.0 & $0.09$\\
CS-R-1: Depth Reasoning (CoT)         & Ego  & 61.6 & \textbf{80.8} & +19.2 & $0.50$\\
CS-R-2: Affordance                   & Ego  & 99.6 & \textbf{100.0} & +0.4 & $0.00$\\
CS-R-3: Causality                    & Ego  & 85.2 & \textbf{93.6} & +8.4 & $0.43$\\
CS-R-4: Exo Spatial$^\star$          & Exo  & 76.6 & \textbf{85.8} & +9.2 & $0.61$\\
\cmidrule{4-6}
\textit{CS Average (6 tasks)} & & \textit{80.9} & \textit{\textbf{91.4}} & \textit{+10.5} & $0.45$\\
\midrule
SP-1: Relative Depth                 & Ego  & 42.0 & \textbf{51.2} & +9.2 & $0.84$\\
SP-2: 360\textdegree{} Layout$^\star$ & Exo & 72.8 & \textbf{97.8} & +25.0 & $0.08$\\
\cmidrule{4-6}
\textit{SP Average (2 tasks)} & & \textit{57.4} & \textit{\textbf{74.5}} & \textit{+17.1} & $0.60$\\
\midrule
IP-1: Arrow-of-Time                  & Ego  & 57.2 & \textbf{83.4} & +26.2 & $0.39$\\
IP-1: Arrow-of-Time$^\star$          & Exo  & 54.0 & \textbf{61.2} & +7.2 & $0.84$\\
IP-2: Object Permanence              & Ego  & 44.0 & \textbf{63.4} & +19.4 & $0.65
$\\
\cmidrule{4-6}
\textit{IP Average (3 tasks)} & & \textit{51.7} & \textit{\textbf{69.3}} & \textit{+17.6} & $0.64$\\
\midrule
MCQ Overlay                           & Mixed & 72.8 & \textbf{95.2} & +22.4 & $0.18$\\
\bottomrule
\end{tabular}%
\vspace*{10mm}
\centering
    \caption{\textbf{Average MCQ accuracy by viewpoint.}}
\label{tab:view_summary}
\begin{tabular}{lccc}
\toprule
\textbf{View} & \textbf{Baseline (\%)} & \textbf{PRISM (\%)} & \textbf{$\Delta$ (\%)} \\
\midrule
Ego (11 tasks) & 57.7 & 83.3 & +25.6 \\
Exo (9 tasks)  & 67.9 & 89.7 & +21.8 \\
Mixed  & 72.8 & 95.2 & +22.4 \\
\midrule
\textbf{Overall } & \textbf{62.8} & \textbf{86.6} & \textbf{+23.8} \\
\bottomrule
\end{tabular}
\end{table*}

\nbf{Embodied Reasoning (ER)}
The largest improvements appear in action reasoning and understanding: ER-3 (goal-conditioned action reasoning) improves by +62.8\%, ER-7 (atomic action CoT) by +55.2\%, and ER-4 (cross-view activity matching) by +48.1\%. Even the challenging multi-actor tasks ER-8 and ER-9, which have no equivalent in Cosmos-Reason training, achieve 94.0\% and 96.4\% respectively. These results demonstrate that domain-specific SFT substantially advances embodied action understanding in retail settings. The experiment also provides an insight into the number of samples required for specific capabilities.

\nbf{Common Sense (CS)}
 Scene understanding CS-U-1, achieves a significant improvement of +21.6\%, indicating robust retail scene understanding. Depth-grounded spatial reasoning CS-R-1 improves by +19.2\%, validating the effectiveness of depth-aware chain-of-thought fine-tuning. This is significant as CS was the best performing category with an avearage accuracy of 80.9\%, which improves to 91.4\%. Even Affordance reasoning CS-R-2, which is already impressive in the baseline, shows a positive improvement as well.

\nbf{Spatial Perception (SP)}
 SP-2, based on the use of the complete 360\textdegree{} layout), achieves an accuracy of 97.8\%, a +25.0\% gain - the largest among spatial tasks - demonstrating that the model learns panoramic spatial reasoning from a previously unseen viewpoint. SP-1 (relative depth) shows a modest +9.2\% gain, reaching 51.2\%. This task requires fine-grained metric depth discrimination that remains challenging even with depth-grounded training data. This could also be a result of the fact that the metric depth was not available to train the original model.

\nbf{Intuitive Physics (IP)}
Among the intuitive physics tasks, IP-1 (arrow of time) shows gains of about +26.2\% and +7.2\%, indicating strong learning of the temporal and causal structure. IP-2 (object permanence) improves by +19.0\%, reflecting better reasoning about the existence of persistent objects under occlusion. Overall, these results suggest that fine-tuning yields meaningful improvements in intuitive physics reasoning.

\nbf{Summary by View}
We also generate performance summaries grouped by camera viewpoint, as shown in \cref{tab:view_summary}. The results indicate that fine-tuning leads to substantial improvements for both egocentric and exocentric tasks, confirming that the model benefits broadly from the multi-view supervision provided by PRISM. Egocentric tasks exhibit slightly larger average gains (+25.6\%) than exocentric tasks (+21.8\%), suggesting that first-person visual streams may provide especially strong signals for learning action-relevant and interaction-centric reasoning. At the same time, the strong gains on exocentric tasks show that the model is also able to leverage third-person scene context effectively. Overall, these trends validate the value of multi-view training data and suggest that combining embodied first-person observations with external scene perspectives leads to more robust viewpoint-generalized reasoning.


\subsection{Data Scaling Analysis}
\label{sec:ablation}
We study how performance scales with training data volume by evaluating checkpoints at 40\%, 60\%, 80\%, and 100\% of the full 270K corpus (\cref{fig:ablation_scaling}). The majority of gains are captured early: at just 40\% of training, overall accuracy reaches 83.5\% (+20.9\% over baseline). Continued training yields diminishing but consistent returns, with the full dataset achieving 88.9\%. Across domains, Embodied Reasoning (ER) shows the most dramatic improvement, jumping from 54.5\% to 86.0\% by the 40\% mark and plateauing near 90.9\%. Common Sense (CS) starts from a strong baseline of 80.9\% and steadily climbs to 91.4\%. Spatial Perception (SP) and Intuitive Physics (IP) improve throughout training but at a slower rate, suggesting these capabilities benefit from broader data coverage. The full-data setting yields the best aggregate performance across all domains.

\input{figs/ablation_scaling}

%% file: figs/ablation_scaling.tex
\begin{figure}[H]
\centering
\begin{tikzpicture}
\begin{axis}[
    width=0.95\columnwidth,
    height=0.7\columnwidth,
    xlabel={\small Training Progress (\% of 270K samples)},
    ylabel={\small MCQ Accuracy (\%)},
    xmin=-5, xmax=108,
    ymin=45, ymax=98,
    xtick={0, 40, 60, 80, 100},
    xticklabels={\textbf{0}, \textbf{40}, \textbf{60}, \textbf{80}, \textbf{100}},
    ytick={50, 55, 60, 65, 70, 75, 80, 85, 90, 95},
    grid=both,
    grid style={gray!15},
    minor grid style={gray!8},
    every axis label/.style={font=\small},
    every tick label/.style={font=\footnotesize},
    axis line style={gray!60},
    tick style={gray!60},
    legend style={
        font=\scriptsize,
        at={(0.02,0.98)},
        anchor=north west,
        draw=gray!40,
        fill=white,
        fill opacity=0.9,
        rounded corners=1pt,
        row sep=1pt,
    },
    legend cell align={left},
]

\addplot[color=black, very thick, line width=1.8pt, smooth, mark=none, forget plot]
    coordinates {(0,62.6)(20,76.0)(40,83.5)(60,87.7)(80,88.5)(100,88.9)};
\addplot[color=black, only marks, mark=*, mark options={fill=black, scale=1.2}]
    coordinates {(0,62.6)(40,83.5)(60,87.7)(80,88.5)(100,88.9)};
\addlegendentry{Overall}

\addplot[color=red!70!black, thick, line width=1.2pt, smooth, mark=none, forget plot]
    coordinates {(0,54.5)(20,74.0)(40,86.0)(60,89.0)(80,90.5)(100,90.9)};
\addplot[color=red!70!black, only marks, mark=square*, mark options={fill=red!70!black, scale=1.0}]
    coordinates {(0,54.5)(40,86.0)(60,89.0)(80,90.5)(100,90.9)};
\addlegendentry{ER (9 tasks)}

\addplot[color=green!50!black, thick, line width=1.2pt, smooth, mark=none, forget plot]
    coordinates {(0,80.9)(20,85.5)(40,88.0)(60,89.5)(80,91.5)(100,91.4)};
\addplot[color=green!50!black, only marks, mark=triangle*, mark options={fill=green!50!black, scale=1.1}]
    coordinates {(0,80.9)(40,88.0)(60,89.5)(80,91.5)(100,91.4)};
\addlegendentry{CS (6 tasks)}

\addplot[color=teal, thick, line width=1.2pt, smooth, mark=none, forget plot]
    coordinates {(0,57.4)(20,63.5)(40,68.0)(60,74.3)(80,75.4)(100,74.5)};
\addplot[color=teal, only marks, mark=diamond*, mark options={fill=teal, scale=1.1}]
    coordinates {(0,57.4)(40,68.0)(60,74.3)(80,75.4)(100,74.5)};
\addlegendentry{SP (2 tasks)}

\addplot[color=blue!70!black, thick, line width=1.2pt, smooth, mark=none, forget plot]
    coordinates {(0,51.7)(20,55.0)(40,58.0)(60,64.0)(80,67.5)(100,69.3)};
\addplot[color=blue!70!black, only marks, mark=pentagon*, mark options={fill=blue!70!black, scale=1.1}]
    coordinates {(0,51.7)(40,58.0)(60,64.0)(80,67.5)(100,69.3)};
\addlegendentry{IP (3 tasks)}

\node[above, font=\tiny, text=green!50!black] at (axis cs:0, 80.9) {80.9};
\node[above, font=\tiny, text=green!50!black] at (axis cs:40, 88.0) {88.0};
\node[above, font=\tiny, text=green!50!black] at (axis cs:60, 89.5) {89.5};
\node[above, font=\tiny, text=green!50!black] at (axis cs:80, 91.5) {91.5};
\node[above, font=\tiny, text=green!50!black] at (axis cs:100, 91.4) {91.4};
\node[below, font=\tiny, text=red!70!black] at (axis cs:0, 54.5) {54.5};
\node[below, font=\tiny, text=red!70!black] at (axis cs:40, 86.0) {86.0};
\node[below, font=\tiny, text=red!70!black] at (axis cs:60, 89.0) {89.0};
\node[below, font=\tiny, text=red!70!black] at (axis cs:80, 90.5) {90.5};
\node[below, font=\tiny, text=red!70!black] at (axis cs:100, 90.9) {90.9};
\node[below, font=\tiny\bfseries] at (axis cs:0, 62.6) {62.6};
\node[below, font=\tiny\bfseries] at (axis cs:40, 83.5) {83.5};
\node[below, font=\tiny\bfseries] at (axis cs:60, 87.7) {87.7};
\node[below, font=\tiny\bfseries] at (axis cs:80, 88.5) {88.5};
\node[below, font=\tiny\bfseries] at (axis cs:100, 88.9) {88.9};
\node[below, font=\tiny, text=teal] at (axis cs:0, 57.4) {57.4};
\node[below, font=\tiny, text=teal] at (axis cs:40, 68.0) {68.0};
\node[below, font=\tiny, text=teal] at (axis cs:60, 74.3) {74.3};
\node[below, font=\tiny, text=teal] at (axis cs:80, 75.4) {75.4};
\node[below, font=\tiny, text=teal] at (axis cs:100, 74.5) {74.5};
\node[below, font=\tiny, text=blue!70!black] at (axis cs:0, 51.7) {51.7};
\node[below, font=\tiny, text=blue!70!black] at (axis cs:40, 58.0) {58.0};
\node[below, font=\tiny, text=blue!70!black] at (axis cs:60, 64.0) {64.0};
\node[below, font=\tiny, text=blue!70!black] at (axis cs:80, 67.5) {67.5};
\node[below, font=\tiny, text=blue!70!black] at (axis cs:100, 69.3) {69.3};

\end{axis}
\end{tikzpicture}
\caption{\textbf{Performance scaling with training data.} MCQ accuracy by capability domain as a function of training progress. All domains improve rapidly in early training (0--40\%), with ER showing the largest absolute gain (+31.5pp). Performance continues to increase at a diminishing rate, with the full 270K dataset achieving the best overall accuracy of 88.9\%. CS starts highest due to strong baseline scene understanding; IP shows steady improvement throughout training.}
\label{fig:ablation_scaling}
\end{figure}
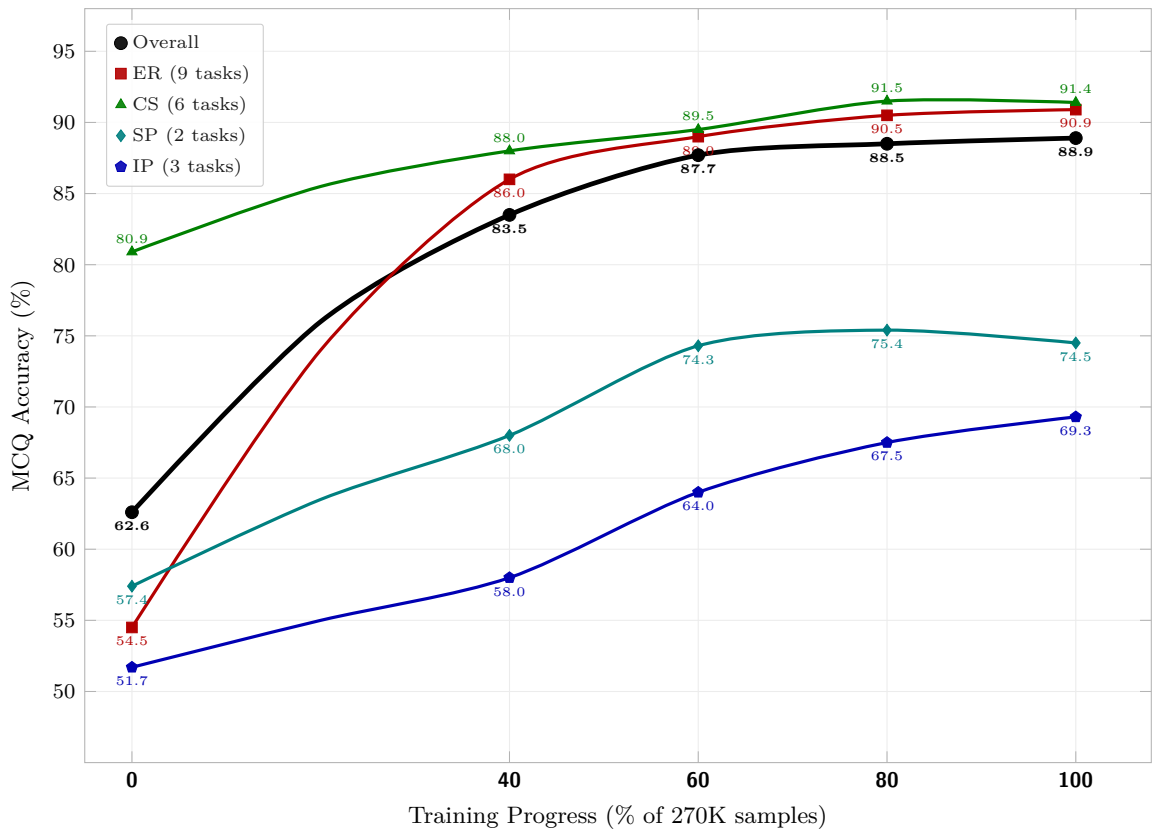

%% file: 10_conclusion.tex
\section{Conclusion}
\label{sec:conclusion}

We presented \textbf{PRISM} — a 270K-sample, multi-view, ontology-structured video SFT corpus for embodied VLMs in real-world retail environments. PRISM is distinguished by three properties that no prior dataset combines: (1) coverage of all three fundamental knowledge dimensions of physical AI—spatial, temporal- physical, and embodied action—within a single deployment domain; (2) synchronized egocentric, exocentric, and 360° panoramic supervision, including eight novel exocentric capability probes with no counterpart in existing corpora; and (3) a scalable, low-cost annotation pipeline combining LLM generation, physics-grounded video reasoning, depth-aware analysis, and self-supervised transformations. Fine-tuning Cosmos-Reason2-2B on PRISM yields a 66.6\% reduction in average error rate across all 20+ probes. Our experimental results establish four actionable conclusions for the embodied AI community. First, domain-specific, ontology-structured SFT dramatically outperforms general pretraining for structured deployment environments: the Embodied Reasoning domain achieves a five-fold error reduction, with individual probes reaching error ratios as low as 0.05. Second, multi-view ego-exo training is mutually reinforcing: exocentric supervision improves cross-view understanding without sacrificing egocentric capability. Third, LLM-generated chain-of-thought annotations substantially outperform template-based alternatives, highlighting that supervision format matters as much as scale. Fourth, the data-scaling curve reveals a favorable property for practitioners: 60\% of the training data captures ~95\% of the total gain, enabling cost-effective partial deployment while the remaining 40\% provides consistent across-the-board improvements.

\nbf{Limitations}
Our study evaluates a single base model (Cosmos-Reason2-2B), limiting conclusions about how PRISM
benefits larger model variants. Automated evaluation metrics (MCQ accuracy, GPT-4o quality scoring)
may underestimate performance on tasks requiring nuanced natural-language outputs.

\nbf{Future Work}
We plan to (1) extend evaluation to larger model variants (7B–13B) to characterize scaling behavior; (2)
incorporate additional sensor modalities (metric depth, inertial measurements) as model inputs for richer spatial grounding; (3) integrate PRISM-trained VLMs into the GR00T \cite{groot_n1} robotic control pipeline for end-to- end evaluation on downstream retail manipulation tasks; and (4) expand data collection to additional geographies and store formats to test the generalization limits of retail-specific SFT.

%% file: 12_appendix.tex
\section{Detailed Capability Examples}
\label{sec:supp_examples}

We provide representative examples from each capability to illustrate the diversity and quality of PRISM samples.

\subsection{ER-1: Next Subtask Prediction}

\begin{table}[H]
\centering
\small
\begin{tabular}{p{0.95\linewidth}}
\toprule
\textbf{Example 1} \\
\midrule
\textbf{Context:} Goal: ``Perform Grocery Shopping''. Sub-goal: ``Navigate Produce Section''. Scene: Produce Section. Completed: ``Aisle Navigation (Fruits and Vegetables).'' \\
\textbf{Q:} What is the next subtask the person will perform? \\
\textbf{A:} The next subtask is: ``Aisle Navigation (Dry Goods)''. Walk forward into an aisle. \\
\midrule
\textbf{Example 2} \\
\midrule
\textbf{Context:} Goal: ``Perform Grocery Shopping''. Sub-goal: ``Select specific spice packets''. Scene: Aisle (Spices section). \\
\textbf{Q:} What is the next subtask the person will perform? \\
\textbf{A:} The next subtask is: ``Item Inspection and Comparison''. \\
\midrule
\textbf{Example 3} \\
\midrule
\textbf{Context:} Goal: ``Perform Grocery Shopping''. Sub-goal: ``Social navigation in produce aisle''. Scene: Produce Section Aisle. \\
\textbf{Q:} What is the next subtask the person will perform? \\
\textbf{A:} The next subtask is: ``Aisle Traversal and Shopper Movement''. The camera continues navigating through the produce section. \\
\bottomrule
\end{tabular}
\end{table}

\subsection{ER-2: Task Completion Verification}

\begin{table}[H]
\centering
\small
\begin{tabular}{p{0.95\linewidth}}
\toprule
\textbf{Example 1} \\
\midrule
\textbf{Q:} Can you determine if the task ``Navigate Aisle while Reviewing List'' has been completed? \\
\textbf{A:} No, the task has not been completed yet. The person walks from the produce section towards the refrigerated freezers, checking their phone. \\
\midrule
\textbf{Example 2} \\
\midrule
\textbf{Q:} Based on what you see, has the person finished ``Wait for Checkout''? \\
\textbf{A:} The task is not yet complete. The person turns back towards the checkout counter, resuming their wait in line. \\
\midrule
\textbf{Example 3} \\
\midrule
\textbf{Q:} Can you determine if the task ``Navigate Aisle'' has been completed? \\
\textbf{A:} The task is not yet complete. The person continues walking forward, navigating between large bags of rice and stacks of cookies. \\
\bottomrule
\end{tabular}
\end{table}

\subsection{ER-3: Goal-Conditioned Action Reasoning}

\begin{table}[H]
\centering
\small
\begin{tabular}{p{0.95\linewidth}}
\toprule
\textbf{Example 1} \\
\midrule
\textbf{Q:} Based on the person's current task, where should they navigate to next in the store? \\
\textbf{A:} $\langle$think$\rangle$The person has just finished inspecting the Heritage Platinum Basmati Rice. Their current sub-goal is ``Examine Product Details,'' which they have completed. The next logical step is to find another product to compare.$\langle$/think$\rangle$ Continue down the current aisle to explore other rice options or adjacent product categories. \\
\midrule
\textbf{Example 2} \\
\midrule
\textbf{Q:} Why is the person performing this action in the context of their shopping goal? \\
\textbf{A:} $\langle$think$\rangle$The person is reaching for a spice packet while their shopping list is visible on their phone. This indicates they are actively checking off items from a prepared list.$\langle$/think$\rangle$ The person is selecting a specific spice that matches an item on their shopping list. \\
\bottomrule
\end{tabular}
\end{table}

\subsection{ER-4: Exo-to-Ego Activity Matching}

\begin{table}[H]
\centering
\small
\begin{tabular}{p{0.95\linewidth}}
\toprule
\textbf{Example 1} \\
\midrule
\textbf{Q:} What activity is the person performing in this exocentric view? \\
\textbf{A:} The worker is scanning shelf labels and restocking items in the beverage aisle. They methodically check each section before placing products from a cart onto the shelves. \\
\midrule
\textbf{Example 2} \\
\midrule
\textbf{Q:} Describe the activity visible from this overhead camera perspective. \\
\textbf{A:} A customer is browsing the produce section, picking up and inspecting individual vegetables before placing selected items into a blue shopping basket. \\
\bottomrule
\end{tabular}
\end{table}

\subsection{ER-5: Hand Interaction Recognition}

\begin{table}[H]
\centering
\small
\begin{tabular}{p{0.95\linewidth}}
\toprule
\textbf{Example 1} \\
\midrule
\textbf{Q:} Describe what the person is doing with their hands in this clip. \\
\textbf{A:} Both hands are holding an item. The person pushes the cart down the aisle, passing shelves filled with rice bags and other packaged goods. \\
\midrule
\textbf{Example 2} \\
\midrule
\textbf{Q:} Describe what the person is doing with their hands in this clip. \\
\textbf{A:} The left hand is holding an item. The right hand is resting and not engaged. The person stops near the vegetable displays and observes another customer selecting carrots from a crate. \\
\bottomrule
\end{tabular}
\end{table}

\subsection{ER-6: Atomic Action Recognition}

\begin{table}[H]
\centering
\small
\begin{tabular}{p{0.95\linewidth}}
\toprule
\textbf{Example 1} \\
\midrule
\textbf{Q:} Based on the video, what action is the person currently performing? \\
\textbf{A:} The action being performed by the man is: reaching toward basket (target: blue shopping basket). Extends right arm towards the blue basket. \\
\midrule
\textbf{Example 2} \\
\midrule
\textbf{Q:} Describe the atomic action visible in this clip. \\
\textbf{A:} The person is performing: walking forward (direction: along aisle). Observable cues: full body translation, upright posture, consistent forward gait. \\
\bottomrule
\end{tabular}
\end{table}

\subsection{ER-7: Atomic Action Reasoning}

\begin{table}[H]
\centering
\small
\begin{tabular}{p{0.95\linewidth}}
\toprule
\textbf{Example 1} \\
\midrule
\textbf{Q:} What is the likely reason this person is doing what they're doing in this scene? \\
\textbf{A:} $\langle$think$\rangle$The staff member is wearing a green vest, which is often an indicator of employee attire. They are bending towards the produce shelf. The action of bending with knees and hips flexing while maintaining a straight back is a biomechanically efficient way to lower one's body to inspect or arrange items on lower shelves.$\langle$/think$\rangle$ The staff member is restocking or inspecting produce items on the lower shelf as part of routine store maintenance. \\
\midrule
\textbf{Example 2} \\
\midrule
\textbf{Q:} Why is this person performing the observed action? \\
\textbf{A:} $\langle$think$\rangle$The customer is standing still near the dairy refrigerator with their head tilted slightly downward, scanning the shelves. This posture suggests they are searching for a specific product among many options.$\langle$/think$\rangle$ The person is comparing product labels to find the right variant or brand of dairy product. \\
\bottomrule
\end{tabular}
\end{table}

\subsection{ER-8: Multi-Actor Scene Understanding}

\begin{table}[H]
\centering
\small
\begin{tabular}{p{0.95\linewidth}}
\toprule
\textbf{Example 1} \\
\midrule
\textbf{Q:} List all the people visible and their current activities in this retail scene. \\
\textbf{A:} The footage captures 2 people. Person~1: A female store employee walks along the produce display, stopping to arrange and organize vegetables on the shelves and bins. Person~2: A man walks forward in the aisle, carrying a blue plastic shopping basket. \\
\midrule
\textbf{Example 2} \\
\midrule
\textbf{Q:} Describe what each person is doing in this scene. \\
\textbf{A:} Three people are visible. Person~1: A customer pushes a cart through the main aisle. Person~2: A staff member arranges products on the end-cap display. Person~3: A shopper examines items on the top shelf, reaching upward. \\
\bottomrule
\end{tabular}
\end{table}

\subsection{ER-9: Social Navigation Reasoning}

\begin{table}[H]
\centering
\small
\begin{tabular}{p{0.95\linewidth}}
\toprule
\textbf{Example 1} \\
\midrule
\textbf{Q:} Describe the actions of each individual visible in this store camera view. \\
\textbf{A:} $\langle$think$\rangle$From this camera view, 4 people can be identified. Person~1 walks into the produce section and interacts with staff. Person~2 is a staff member assisting at the vegetable display. Person~3 browses independently near the refrigerated section. Person~4 navigates toward the checkout area.$\langle$/think$\rangle$ Four individuals occupy the store with distinct goals: one customer selects produce with staff assistance, another browses independently, and a fourth heads toward checkout. \\
\bottomrule
\end{tabular}
\end{table}

\subsection{CS-U-1: Scene Description VQA}

\begin{table}[H]
\centering
\small
\begin{tabular}{p{0.95\linewidth}}
\toprule
\textbf{Example 1} \\
\midrule
\textbf{Q:} Describe what is happening in this scene. \\
\textbf{A:} The person stands at the open refrigerator door, looking at the milk pouches and curd packages inside. The camera captures the person's hands near the shelves. Another person walks past in the background. \\
\midrule
\textbf{Example 2} \\
\midrule
\textbf{Q:} Describe the environment and surroundings in this scene. \\
\textbf{A:} This is the Aisle (Sweets/Snacks) section of the supermarket. The person's current sub-goal is: Examine Item Details and Check Shopping List. \\
\bottomrule
\end{tabular}
\end{table}

\subsection{CS-U-2: Environment Understanding VQA}

\begin{table}[H]
\centering
\small
\begin{tabular}{p{0.95\linewidth}}
\toprule
\textbf{Example 1} \\
\midrule
\textbf{Q:} What is the typical arrangement of products on the shelving units in this grocery store aisle? \\
\textbf{A:} The shelving units display a variety of packaged food items, primarily snacks, arranged by category. Products are neatly stocked and often stacked to maximize shelf space, with labels facing forward for easy identification by customers. \\
\midrule
\textbf{Example 2} \\
\midrule
\textbf{Q:} Describe the layout of the produce section from this exocentric view. \\
\textbf{A:} The produce section features open bins along both sides of a central walkway, with fruits on the left and vegetables on the right. Overhead signage indicates product categories. A weighing scale is positioned at the end of the display. \\
\bottomrule
\end{tabular}
\end{table}

\subsection{CS-R-1: Spatial Reasoning with Chain-of-Thought}

\begin{table}[H]
\centering
\small
\begin{tabular}{p{0.95\linewidth}}
\toprule
\textbf{Example 1} \\
\midrule
\textbf{Q:} Are the visible items generally close to the person, or spread across a wider range? \\
\textbf{A:} $\langle$think$\rangle$The depth statistics show a mean depth of 0.172. This is a relatively low value, indicating that most objects are in the closer to mid-range. The minimum depth is 0.055, suggesting something very close like the basket on the floor...$\langle$/think$\rangle$ Items are primarily close-range. \\
\midrule
\textbf{Example 2} \\
\midrule
\textbf{Q:} What is the shopper's current state, and what does this imply for robot navigation? \\
\textbf{A:} $\langle$think$\rangle$I observe a human shopper in the aisle in a ``Dynamic'' state. This dynamism is evidenced by the shopper actively picking up an item from a shelf.$\langle$/think$\rangle$ The shopper is dynamic; the robot should maintain safe distance and anticipate movement. \\
\bottomrule
\end{tabular}
\end{table}

\subsection{CS-R-2: Affordance Reasoning}

\begin{table}[H]
\centering
\small
\begin{tabular}{p{0.95\linewidth}}
\toprule
\textbf{Example 1} \\
\midrule
\textbf{Q:} Based on the spatial layout, what affordances are available to the person? \\
\textbf{A:} $\langle$think$\rangle$The shelves on both sides are within immediate reaching distance, affording grasping of products like snacks, beverages, and household items. The floor affords walking through the aisle. The shopping basket provides containment for selected items.$\langle$/think$\rangle$ The environment affords navigation through aisles and grasping items from shelves and bins. The basket facilitates collection of products. \\
\bottomrule
\end{tabular}
\end{table}

\subsection{CS-R-3: Causality Reasoning}

\begin{table}[H]
\centering
\small
\begin{tabular}{p{0.95\linewidth}}
\toprule
\textbf{Example 1} \\
\midrule
\textbf{Q:} Analyze the cause-and-effect: why did this action happen? \\
\textbf{A:} $\langle$think$\rangle$The person is in an aisle stocked with dry goods and beverages. Their goal is to perform grocery shopping, which necessitates navigating aisles to find items on their list. Walking down this aisle directly progresses through the store's layout toward the target items.$\langle$/think$\rangle$ To find items on their shopping list located in the dry goods and beverages aisle. \\
\bottomrule
\end{tabular}
\end{table}

\subsection{CS-R-4: Spatial Reasoning (Exocentric)}

\begin{table}[H]
\centering
\small
\begin{tabular}{p{0.95\linewidth}}
\toprule
\textbf{Example 1} \\
\midrule
\textbf{Q:} Describe the typical person-object interaction required to select items from the bulk bins and refrigerated cases. \\
\textbf{A:} $\langle$think$\rangle$To interact with the bulk bins, a customer would reach over the counter, lift the transparent lid, and use a scoop to retrieve the product. For refrigerated cases, the customer would open one of the glass doors to access and select items.$\langle$/think$\rangle$ Customers interact with bulk bins by lifting lids and scooping products, and with refrigerated cases by opening glass doors to retrieve items. \\
\bottomrule
\end{tabular}
\end{table}

\subsection{SP-1: Relative Depth Reasoning}

\begin{table}[H]
\centering
\small
\begin{tabular}{p{0.95\linewidth}}
\toprule
\textbf{Example 1} \\
\midrule
\textbf{Q:} Does this scene have a wide or narrow depth range? \\
\textbf{A:} This scene has a moderate depth range from 0.05 to 0.78. There is a mix of closer and farther objects, but the distances do not span an extreme range. \\
\midrule
\textbf{Example 2} \\
\midrule
\textbf{Q:} Are objects in the top portion of the view closer or farther than those at the bottom? \\
\textbf{A:} Objects in the top portion of the view are at shorter distances than those at the bottom, indicating a downward-looking perspective toward the floor and nearby shelves. \\
\bottomrule
\end{tabular}
\end{table}

\subsection{SP-2: 360\textdegree{} Spatial Layout Reasoning}

\begin{table}[H]
\centering
\small
\begin{tabular}{p{0.95\linewidth}}
\toprule
\textbf{Example 1} \\
\midrule
\textbf{Q:} Describe the spatial layout and obstacle proximity from this panoramic view. \\
\textbf{A:} $\langle$think$\rangle$Mean depth 0.43, bimodality 0.07. The panoramic view reveals a central open aisle flanked by shelving units on both sides at moderate distance.$\langle$/think$\rangle$ Open central aisle, shelves on both sides at moderate distance. No immediate obstacles in the navigation corridor. \\
\bottomrule
\end{tabular}
\end{table}

\subsection{IP-1: Arrow-of-Time}

\begin{table}[H]
\centering
\small
\begin{tabular}{p{0.95\linewidth}}
\toprule
\textbf{Example 1 (Forward)} \\
\midrule
\textbf{Q:} Determine whether this video is playing in its original direction or in reverse. \\
\textbf{A:} This clip is playing forward. The person walks down a wide aisle, with shelves displaying biscuits and other packaged snacks on both sides. \\
\midrule
\textbf{Example 2 (Backward)} \\
\midrule
\textbf{Q:} Is this video playing forward or backward? \\
\textbf{A:} The video has been reversed and is playing backward. The person places the selected ghee container into the shopping trolley. \\
\bottomrule
\end{tabular}
\end{table}

\subsection{IP-1 CoT: Physics Reasoning}

\begin{table}[H]
\centering
\small
\begin{tabular}{p{0.95\linewidth}}
\toprule
\textbf{Example 1 (Ego, Backward)} \\
\midrule
\textbf{Q:} Watch this video and determine if it is playing forward or backward. Analyze the body mechanics and biomechanical cues. \\
\textbf{A:} $\langle$think$\rangle$While the walking motion appears smooth, the foot placement and overall gait reveal subtle inconsistencies. The natural push-off and weight transfer of a forward stride are absent, giving the movement a slightly unnatural, gliding quality.$\langle$/think$\rangle$ The video is playing backward. \\
\midrule
\textbf{Example 2 (Exo, Forward)} \\
\midrule
\textbf{Q:} Watch this video and determine if it is playing forward or backward. Analyze the temporal flow and causal relationships. \\
\textbf{A:} $\langle$think$\rangle$A person enters the frame from the left and walks toward the right. The legs move in a natural forward progression. There are no visual cues of reverse motion or unnatural movement patterns.$\langle$/think$\rangle$ The video is playing forward. \\
\bottomrule
\end{tabular}
\end{table}

\subsection{IP-2: Object Permanence}

\begin{table}[H]
\centering
\small
\begin{tabular}{p{0.95\linewidth}}
\toprule
\textbf{Example 1} \\
\midrule
\textbf{Q:} Something left the field of view. Does the object continue to exist when not visible? \\
\textbf{A:} The object still exists. In a retail environment, items on shelves, carts, and people maintain their physical presence even when they leave the camera frame. Object permanence is a basic physical principle. \\
\midrule
\textbf{Example 2} \\
\midrule
\textbf{Q:} A person walked behind a shelf and is no longer visible. Do they still exist? \\
\textbf{A:} Yes, the person continues to exist behind the shelf. Occlusion by a physical barrier does not cause objects or people to cease existing. They will likely reappear when the viewpoint changes or they move past the obstruction. \\
\bottomrule
\end{tabular}
\end{table}

\subsection{MCQ Overlay}

\begin{table}[H]
\centering
\small
\begin{tabular}{p{0.95\linewidth}}
\toprule
\textbf{Example 1} \\
\midrule
\textbf{Q:} What hand activity can you observe in this video? \\
A.~One hand is actively engaged in gesturing while the other is still. \\
B.~Both hands are idle and resting at the person's sides. \\
C.~The person is holding an object in their right hand and gesturing with the left. \\
D.~The person is using both hands to clap rhythmically. \\
\textbf{A:} B \\
\bottomrule
\end{tabular}
\end{table}

\subsection{MCQ Standalone}

\begin{table}[H]
\centering
\small
\begin{tabular}{p{0.95\linewidth}}
\toprule
\textbf{Example 1 (Physics)} \\
\midrule
\textbf{Q:} When a shopping cart is pushed and then released on a flat store floor, it gradually slows down and stops. What is the primary reason? \\
A.~The cart runs out of energy. \\
B.~Friction between the wheels and the floor dissipates kinetic energy. \\
C.~Gravity pulls the cart downward. \\
D.~Air resistance is the dominant force. \\
\textbf{A:} B \\
\midrule
\textbf{Example 2 (Retail)} \\
\midrule
\textbf{Q:} In a typical grocery store, which section is most likely located near the entrance? \\
A.~Frozen foods. \\
B.~Fresh produce. \\
C.~Cleaning supplies. \\
D.~Canned goods. \\
\textbf{A:} B \\
\bottomrule
\end{tabular}
\end{table}

\clearpage
\section{Fine-Tuning Details}
\label{sec:supp_finetuning}

\subsection{Fine-Tuning Configuration}

\begin{table}[H]
\centering
\caption{\textbf{Fine-tuning hyperparameters.}}
\label{tab:supp_qlora}
\begin{tabular}{ll}
\toprule
\textbf{Parameter} & \textbf{Value} \\
\midrule
Base model & Cosmos-Reason2-2B \\
Architecture & Qwen3-VL \\
Quantization & None (full BF16) \\
Compute dtype & bfloat16 \\
LoRA rank ($r$) & 32 \\
LoRA alpha ($\alpha$) & 64 \\
LoRA target & q,k,v,o,gate,up,down\_proj (LM only) \\
Total params & 2.49B \\
\midrule
Optimizer & AdamW \\
Learning rate & $1 \times 10^{-4}$ \\
LR schedule & Cosine \\
Warmup & 5\% of total steps (397 steps) \\
Batch size & 1 per GPU \\
Gradient accumulation & 8 \\
Effective batch size & 32 \\
Max sequence length & 4,096 \\
Max pixels & 360,000 \\
Epochs & 1 \\
Gradient checkpointing & Enabled (non-reentrant) \\
Attention & SDPA \\
\midrule
Hardware & 4$\times$ RTX PRO 6000 Blackwell (98\,GB) \\
Training time & $\sim$35 hours (7,942 steps) \\
\bottomrule
\end{tabular}
\end{table}


\subsection{Qwen3-VL Temporal Frame Handling}

During implementation, we discovered a compatibility issue between the Qwen3-VL processor and the \texttt{get\_rope\_index} method when processing video inputs. The processor creates per-frame token groups in \texttt{mm\_token\_type\_ids} separated by timestamp tokens, but \texttt{video\_grid\_thw} contains a single $[T, H, W]$ entry. The \texttt{get\_rope\_index} method expects one grid entry per contiguous type-2 group, causing a \texttt{StopIteration} error.

Our solution monkey-patches \texttt{get\_rope\_index} to internally split each $[T, H, W]$ entry into $T$ rows of $[1, H, W]$, matching the per-frame token groups. The visual encoder receives the original unsplit grid for correct spatial processing. This fix is necessary for both training and inference with video inputs.




\section{Contributions and Acknowledgments}
\label{sec:contribution}

\subsection{Core Contributors: }
Amirreza Rouhi, Parikshit Sakurikar, Satya Sai Reddy, Narsimha Menga, Anirudh Govil, Sri Harsha Chittajallu, Rajat Aggarwal, Anoop Namboodiri, and Sashi Reddi.

\subsection{Acknowledgments: }
The authors express their sincere gratitude to Deepu Tiwari, Suresh K, Shrinivas Gone, Sirigiri Manikanta, Farooq Basha Chowdary, Bagara Sai Varun, Yedida Phani Sri Sathvik, Boddu Shyamala, Shanmuki Priya Bala, Nasika Lokesh Vara Prasad, Sukanya Marri, Mahajan Manikanta, Pasunoori Venkatsai, Manisha, K Uday Kiran, Induri Srikanth Reddy, Ayapakola Dwarkesh, Nihasri Gundu, Cholleti Vinay Kumar, Rudra. Saikiran, Akuthota Yashwanth, Achintalwar Divya Jyothi, Akkala Varshitha, Gadthy Sai Keerthika, Shaik Mohammad Irfan, C Bharath, Gorli Revathi for their valuable assistance with data collection and annotation.